\documentclass{article}

\usepackage[main, final]{neurips_2026}

\usepackage{wrapfig}
\usepackage{graphicx}
\usepackage{amsmath}
\newtheorem{definition}{Definition}

\usepackage[utf8]{inputenc} % allow utf-8 input
\usepackage[T1]{fontenc}    % use 8-bit T1 fonts
\usepackage{hyperref}       % hyperlinks
\usepackage{url}            % simple URL typesetting
\usepackage{booktabs}       % professional-quality tables
\usepackage{amsfonts}       % blackboard math symbols
\usepackage{nicefrac}       % compact symbols for 1/2, etc.
\usepackage{microtype}      % microtypography
\usepackage{xcolor}         % colors
\usepackage{subcaption}
\usepackage{adjustbox}
\title{Compressible Representations: Functional Spines in Deep Neural Networks}

\author{%
  Niranjan Rajesh, Meenakshi Khosla\\
  Department of Cognitive Science\\
  University of California San Diego\\
  La Jolla, CA 92092 \\
  \texttt{\{nirajesh, mkhosla\}@ucsd.edu} \\
}

\begin{document}

\maketitle

\renewcommand{\thefootnote}{}
\footnotetext{All code is publicly available at:
\url{https://github.com/NeuroML-Lab/compressible-representations}}
\addtocounter{footnote}{-1}

% \footnotetext[]{All code is publicly available at: 
% \url{https://github.com/NeuroML-Lab/compressible-representations}}
\begin{abstract}
Deep networks embed inputs in high-dimensional representations, but how much of this capacity actually drives behavior? We introduce a framework for identifying,
at each layer, the subspace of the representation that is functionally sensitive to the task. Using a matrix analogous to the Fisher information but defined over the hidden representation, we rank directions by their influence on downstream computation
and project onto the most task-sensitive subspace. We benchmark this behaviorally
grounded compression against three geometric alternatives that preserve variance,
local neighborhood structure, or global topology. We investigate the utility of the task-sensitive subspace across a range of domains concerned with network representations. Across pretrained vision networks,
representations are remarkably compressible: a small fraction of the task-sensitive
subspace preserves most performance, often requiring far fewer dimensions than
geometric alternatives. We find that this compressed representation is also more robust to adversarial attacks generated on the full network. We additionally show that this decomposition recovers functionally
distinct subspaces within the same representation: on cue-conflict stimuli,
shape-sensitive and texture-sensitive directions are largely non-overlapping, and
ablating one selectively impairs the corresponding task while leaving the other
intact. Task-sensitive subspaces also provide a principled foundation for continual
learning: constraining gradient updates to lie orthogonal to the task-sensitive
subspaces of prior tasks outperforms standard variance-based gradient projection
methods. Finally, restricting representational comparisons to task-sensitive
subspaces reveals that inter-model alignment is higher in task-sensitive directions
than in task-insensitive directions at later layers, and that conditioning on different
tasks over the same stimuli uncovers asymmetries that full-space similarity
measures cannot detect. Together, these results establish that deep networks
compute through a low-dimensional functional spine embedded within a much
larger representational space, that these spines differ across tasks, and that they provide a unifying basis for compression, continual learning, and representational comparison. 
\end{abstract}

\section{Introduction}
Deep networks are vastly overparameterized in their weight space. The lottery ticket hypothesis \citep{frankle2018lottery} demonstrated that sparse subnetworks can match the performance of full networks; knowledge distillation \citep{hinton2015distilling} showed that the functional behavior of large networks can be captured by much smaller ones; and extensive work on pruning and quantization \citep{han2015deep, galeStateSparsityDeep2019, gholami2022survey} has established that most parameters can be removed with minimal impact on task performance. This line of work has produced both practical tools for model compression and theoretical insights into the structure of loss landscapes and generalization.

But overparameterization in weights does not directly imply overparameterization in representations. The question of how much representational capacity a network actually uses for its task has received comparatively less attention.
%A network might require all its parameters to produce hidden states that are themselves low-dimensional, or conversely, a heavily prunable network might produce representations that fully occupy their ambient space. 

There are reasons to expect that representations are compressible. The intrinsic dimensionality of learned representations has been studied through the lens of manifold geometry and several works have estimated that neural representations lie on manifolds of much lower dimension than the ambient space \citep{ansuini2019intrinsic}. However, intrinsic dimensionality is a geometric property - it characterizes the shape of the data manifold but does not distinguish between directions that matter for behavior and directions that do not. 

%A representation could lie on a low-dimensional manifold and yet require all dimensions of that manifold for the task, or it could occupy a high-dimensional manifold of which only a small subspace is behaviorally relevant. Characterizing which directions of the representational space are functionally relevant requires a criterion derived from the network's downstream computation, not in the geometry alone.

A separate line of work has studied functional importance through probing \citep{alain2016understanding} and causal intervention \citep{ferrando2024primer}. Linear probes test whether task-sensitive information is linearly decodable from a representation, but high probe accuracy does not imply that the network itself uses that information downstream \citep{ravichander2021probing}. Causal approaches, including activation patching \citep{vig2020investigating}, causal tracing \citep{meng2022locating}, and interchange interventions \citep{geiger2022inducing}, directly test whether specific components of the representation influence behavior, but typically operate at the level of individual units or attention heads rather than empirically identifying continuous subspaces.

The question of representational comparison faces a related challenge. Methods such as CKA \citep{kornblith2019similarity}, RSA \citep{kriegeskorte2008representational}, and shape distances \citep{williams2021generalized} compare representations by measuring similarity in the full ambient space. They are often dominated by directions of high variance, regardless of whether those directions are sensitive to behavior. If most dimensionality is behaviorally inert, standard comparisons may be driven by structure that plays no role in the network's computation. Indeed, several recent works have noted that representational similarity measures can disagree with each other and with functional measures of similarity \citep{ding2021grounding, bo2024evaluating, cloos2024differentiable, klabunde2305similarity, braun2025not}, raising the question of whether comparisons should be restricted to the subspace that matters.

In this work, we study representational compressibility directly in deep vision networks. Using a matrix analogous to the Fisher information but defined over hidden representations, we identify the subspace along which downstream computation is most concentrated -- the functional spine of the network. We benchmark this against three geometric alternatives (PCA, local-neighborhood, geotopology) and find that our task-sensitive compression substantially outperforms geometric alternatives at middle and later layers. We further show that the decomposition recovers functionally distinct (shape vs texture) and adversarially robust subspaces, improves continual learning when used to constrain gradient updates, and enables task-conditioned representational comparisons that reveal inter-network asymmetries invisible to full-space similarity measures.

\section{Methods}

We study the \emph{compressibility} of a representation by asking whether a hidden state
$z \in \mathbb{R}^d$ can be replaced by a lower-dimensional projection
without substantially degrading downstream behavior.
Concretely, we consider interventions of the form $\tilde z = \Pi_r z = U_r U_r^\top z$, where $U_r \in \mathbb{R}^{d \times r}$ has orthonormal columns and $r<d$.
Although the representation lives in a $d$-dimensional ambient space, the downstream computation may depend only on a much smaller set of effective degrees of freedom.
Representational compressibility is therefore the extent to which functional performance can be preserved when communication through the representation is restricted to a low-rank subspace.

\paragraph{Task-sensitive compressibility}
Our primary notion of compressibility is grounded purely in the
functional role of the representation for downstream computation.
Let $g: \mathbb{R}^{n_\ell} \to \mathbb{R}^k$ denote the mapping
from the hidden state $z_\ell$ at layer $\ell$ (with $n_\ell$ units) through all
subsequent layers to a downstream output representation
(e.g., class logits for $k$ classes), and let $J^g_\ell(x) = \frac{\partial g}{\partial z_\ell}(x)
\in \mathbb{R}^{k \times n_\ell}$ be its Jacobian evaluated at input $x$.
\begin{definition}[Task-sensitivity matrix]
The task-sensitivity matrix at layer $\ell$ is
\begin{equation}
F^g_\ell = \frac{1}{M}\sum_{i=1}^{M} J^g_\ell(x_i)^\top J^g_\ell(x_i)
\in \mathbb{R}^{n_\ell \times n_\ell}.
\label{eq:F}
\end{equation}
\end{definition}
\noindent
This positive semidefinite matrix measures how sensitive the downstream
output is to perturbations in each direction of the activation space,
averaged over the data with $M$ samples. For any unit vector $v$, the quadratic form
$v^\top F^g_\ell v$ gives the mean squared change in downstream output
induced by a unit perturbation along $v$. Thus, $F^g_\ell$ is the
pullback of the Euclidean metric on output space back to the
representation space at layer $\ell$. The task-sensitive rank-$r$ projector is
$\Pi^g_r = V_{:,1:r}V_{:,1:r}^\top$, i.e., the projector onto the top
$r$ eigenvectors of $F^g_\ell$.

% \begin{definition}[Task-sensitive and null subspaces]
% Let $F^g_\ell = V \Lambda V^\top$ with eigenvalues
% $\lambda_1 \ge \lambda_2 \ge \cdots \ge \lambda_{n_\ell} \ge 0$.
% The task-relevant rank-$r$ projector is
% $\Pi^g_r = V_{:,1:r}V_{:,1:r}^\top$, i.e., the projector onto the top
% $r$ eigenvectors of $F^g_\ell$. The corresponding null-space projector is
% $\Pi^{g,\perp}_r = I - \Pi^g_r$, which projects onto the remaining
% eigenvectors.
% \end{definition}

\noindent
Projecting hidden states with $\Pi^g_r$ retains only the directions
along which perturbations most affect downstream behavior, while
projecting with $\Pi^{g,\perp}_r$ isolates the complementary
subspace (or the \textit{null}-space) that is the least sensitive to task performance. 

This perspective is analogous to Fisher-information-based analyses of perceptual discriminability \citep{berardino2017eigen, feather2024discriminating}, but applied to representation space instead of the stimulus space: rather than asking which stimulus perturbations most change neural responses, we ask which representational perturbations most change downstream behavior.

\paragraph{Variance-preserving compressibility.} A second notion of compressibility is purely geometric.
Given a matrix of representations $Z_\ell \in \mathbb{R}^{n \times d}$ with centered version $Z_{\ell,c}$, Principal Components Analysis (PCA) defines a low-rank subspace from the covariance $C_\ell = \frac{1}{n} Z_{\ell,c}^\top Z_{\ell,c}$.
Let $C_\ell = V \Lambda V^\top$ be the eigendecomposition with eigenvalues sorted in descending order. The rank-$r$ variance-preserving projector is $\Pi_r^{\mathrm{PCA}} = V_r V_r^\top$, where $V_r$ contains the top $r$ eigenvectors.
Projecting with $\Pi_r^{\mathrm{PCA}}$ retains the directions that explain the greatest variance in the representation. This notion of compressibility is also closely related to preserving global Euclidean geometry. See Appendix \ref{app:more_methods}1 for more detail.

\paragraph{Local-neighborhood-preserving compressibility.}
A third notion of compressibility focuses on local rather than global geometry.
Instead of preserving overall variance or full pairwise distance structure, we ask which directions are needed to preserve the neighborhood relations among nearby stimuli.
Let $W \in \mathbb{R}^{n \times n}$ be a symmetric weight matrix constructed from a $k$-nearest-neighbor graph over the representations, where $W_{ij} > 0$ if stimulus $j$ is among the neighbors of stimulus $i$ (or vice versa), and $W_{ij} = 0$ otherwise.
Let $L = D - W$ denote the graph Laplacian, where $D$ is the diagonal degree matrix with $D_{ii} = \sum_j W_{ij}$.
The local-geometry matrix is $M_\ell^{\mathrm{local}} = 2\, Z_{\ell}^\top L\, Z_{\ell}$.
For any direction $v$, the quadratic form $v^\top M_\ell^{\mathrm{local}}\, v = \sum_{i,j} W_{ij}(v^\top(z_i - z_j))^2$ measures how much of the local neighbor-difference structure is captured along $v$.
The top eigenvectors of $M_\ell^{\mathrm{local}}$ therefore define the subspace that best preserves the local manifold geometry of the representation.
If $M_\ell^{\mathrm{local}} = Q \Gamma Q^\top$, the rank-$r$ local-neighborhood projector is $\Pi_r^{\mathrm{local}} = Q_r Q_r^\top$, where $Q_r$ contains the top $r$ eigenvectors.

\paragraph{Geotopology-preserving compressibility.}
A fourth notion of compressibility focuses on preserving a geotopological (GT) summary of the representation that emphasizes intermediate-range relational structure.
This notion is motivated by the framework of \citet{lin2024topology}, who introduced a family of nonlinear monotonic transforms of representational distances that balance sensitivity to geometry and topology.
The key insight is that very small distances are noise-dominated and very large distances are idiosyncratic across network instances, so the most informative variation lies in an intermediate band. For a technical explanation of the GT transform, please refer to Appendix \ref{app:more_methods}2.

% The GT transform induces a qualitatively different weighting over stimulus pairs than the other three compression criteria.
% PCA weights all pairwise distances equally through global variance.
% Local-neighborhood preservation assigns nonzero weight only to nearby pairs.
% The GT transform inverts this emphasis: 
The GT transform assigns \emph{zero} weight to the closest pairs (below $l$), graded weight in an intermediate band (between $l$ and $u$), and \emph{constant} weight to the most distant pairs (above $u$).
It therefore functions as a band-pass filter on representational geometry, preserving the directions that capture variation in the intermediate distance range where \citet{lin2024topology} argue the most functionally informative structure resides. Task-sensitivity, by contrast, is orthogonal to all three geometric criteria, weighting directions by their causal influence on downstream computation rather than by any property of the distance distribution.

These four projections formalize distinct notions of representational compressibility (behavioral sensitivity, global variance, local geometry, and mesoscale relational structure) none of which have previously been used as interventional compression criteria applied mid-network. A key distinction is that the task-sensitivity matrix requires backpropagation through the downstream network and is sensitive to estimation quality, while the geometric criteria are computed directly from the representation. Whether these notions agree or diverge, and at which layers, is an empirical question. 

\paragraph{Random projections.}
Finally, as a baseline, we also evaluate compression onto a random subspace of the same rank $r$. We construct a random orthonormal basis by eigendecomposing a random symmetric matrix and retaining random $r$ eigenvectors, which is equivalent to selecting a uniformly random $r$-dimensional subspace. Random projections preserve no particular structure (geometric or behavioral) and therefore provide a lower bound against which all structured compression criteria can be compared.

%%%TODO
% \paragraph{Global compression.}
% The criteria above compress one layer at a time while leaving the rest of the network unchanged, so they do not show whether compression effects accumulate or interact across layers. We therefore also compress all $L$ layers simultaneously under a fixed total rank budget $r_{\text{total}} = \lfloor \rho \sum_{\ell=1}^{L} n_\ell \rfloor$, where $n_\ell$ is the dimensionality of the representation at layer $\ell$ and $\rho \in (0,1)$ is the global retained fraction. The per-layer ranks $r_\ell$ are chosen in one of two ways. \emph{Fixed allocation} retains the same fraction of dimensions at every layer, i.e., $r_\ell = \lfloor \rho \, n_\ell \rfloor$. \emph{Dynamic allocation} is performance-aware: we define the utility $u_\ell(r)$ as the top-5 accuracy obtained by compressing only layer $\ell$ to rank $r$ (from the single-layer curves), and use dynamic programming to solve $\max_{\{r_\ell\}} \sum_{\ell=1}^{L} u_\ell(r_\ell)$ subject to $\sum_{\ell} r_\ell \le r_{\text{total}}$. Since compressibility varies with depth, this assigns more dimensions to layers where compression is costly and fewer to layers where it is cheap. Together, these experiments test whether layer-wise compression trends compose into effective end-to-end compression.
We evaluate all compression methods both layer-wise and globally. In the layer-wise setting, a single layer is compressed while all others are left unmodified. In the global setting, all layers are compressed simultaneously under a fixed total rank budget, expressed as a fraction of the total dimensionality across layers. The budget is distributed either uniformly, retaining the same fraction of dimensions at every layer, or dynamically, with a performance-aware allocation that assigns more dimensions to layers where compression is costly and fewer to layers where it is cheap (Appendix~\ref{app:accuracy_eval}). Together, these settings test whether compression effects hold when isolated in a single layer and whether they accumulate or interact across the network.
% In practice, the retained rank $r$ may be
% chosen either by a single-layer criterion or by a global utility-based
% allocation under a fixed compression budget for any of the compression types. 

% The layer-wise experiments compress one layer at a time while leaving the rest of the network unchanged. They therefore do not show whether compression effects accumulate or interact when several layers are compressed simultaneously. The global retention experiment addresses this by allocating a fixed total dimensionality budget across layers and applying all selected projections at once. We compare a uniform allocation with a performance-aware allocation derived from the single-layer compression curves. Overall this tests whether layer-wise compression trends compose into effective end-to-end compression and whether they remain robust under a more substantial intervention. We will clarify this purpose in the revision

\section{Results}
\begin{wrapfigure}{r}{0.43\linewidth}
  \vspace{-1em}
  \centering
  \includegraphics[width=0.99\linewidth]{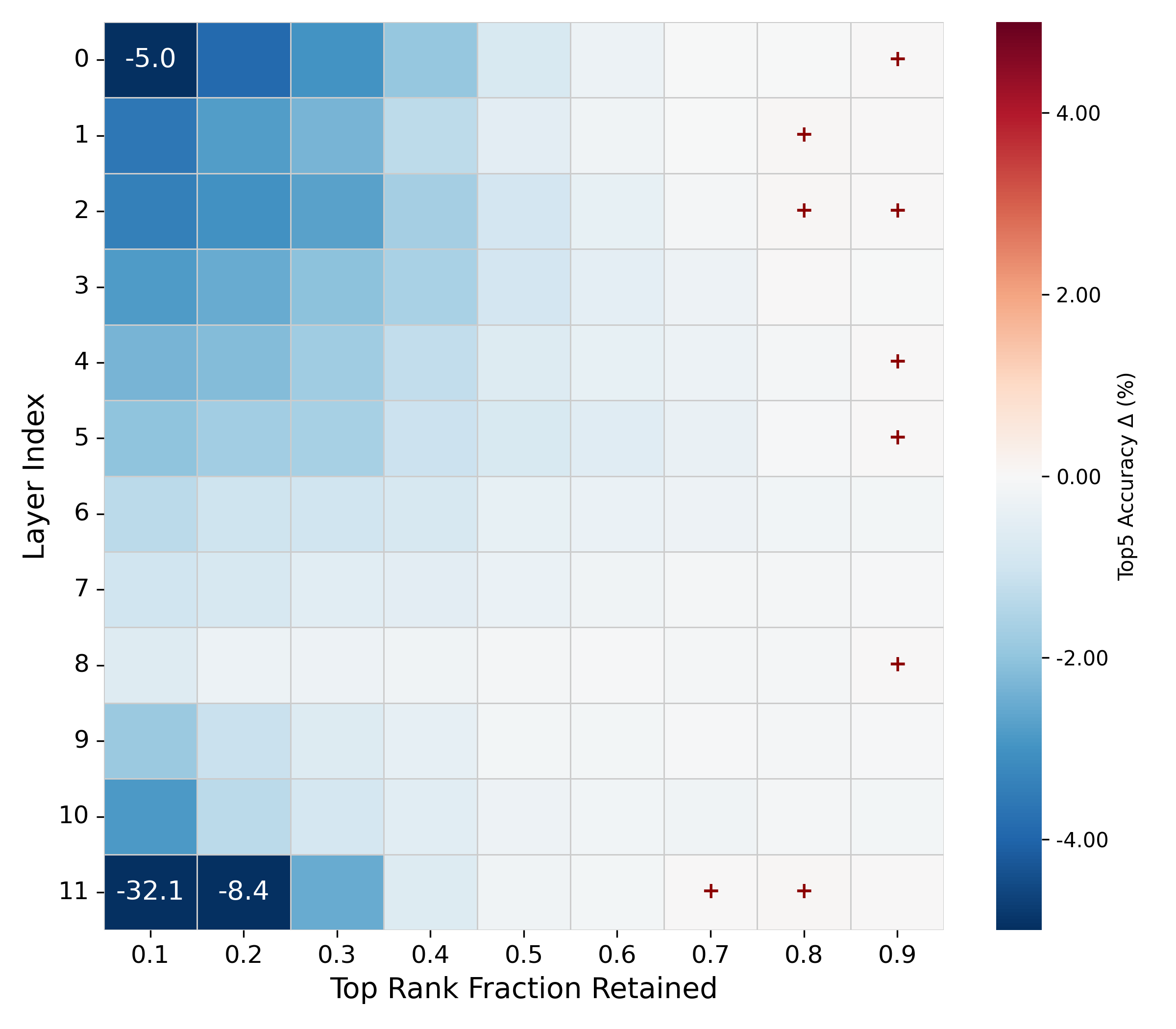}
  \caption{
    \textbf{Representations are dramatically compressible along task-sensitive directions.} Top-5 ImageNet accuracy drop from baseline when projecting onto the top-$r$\% task-sensitive directions at each layer. $+$: accuracy exceeds uncompressed baseline.}
  \label{fig:a1_task_relevance_heatmap}
  \vspace{-1em}
\end{wrapfigure}

\subsection{Task-sensitive subspaces preserve performance under aggressive compression}
We begin by asking a basic question: how much of each layer's representation is functionally necessary for the network's task performance? For each layer of a pretrained ViT-B/16 (baseline top-5: 95.1\%), we compute the task-sensitivity matrix $F^g_\ell$ (Eq.~\ref{eq:F}), project hidden states onto the top-$r$ eigenvectors, and evaluate ImageNet classification accuracy.

Figure~\ref{fig:a1_task_relevance_heatmap} shows the difference in top-5 accuracy between the compressed network and the base network as a function of layer and retained rank fraction for task-sensitive projections.
Representations are remarkably compressible: at middle layers, retaining a small fraction of the task-sensitive subspace preserves most of the network's performance.
For example, at layer~7, projecting the 768-dimensional representation onto its top 77 task-sensitive directions (just 10\% of the full rank) preserves 94.1\% top-5 accuracy (98.9\% of the uncompressed baseline). In some cases, moderate compression even marginally exceeds the baseline, likely because projecting out low-sensitivity directions removes variation that slightly interferes with downstream computation.

\subsection{Task-sensitive compression outperforms geometric alternatives} 
Does compressibility require a task-grounded criterion, or do standard geometric projections suffice?
We compare task-sensitive compression against three alternatives, including PCA, local-neighborhood preservation, and geotopology preservation, across all layers and compression levels.

\begin{figure}[htbp]
    \centering
    \includegraphics[width=0.8\linewidth]{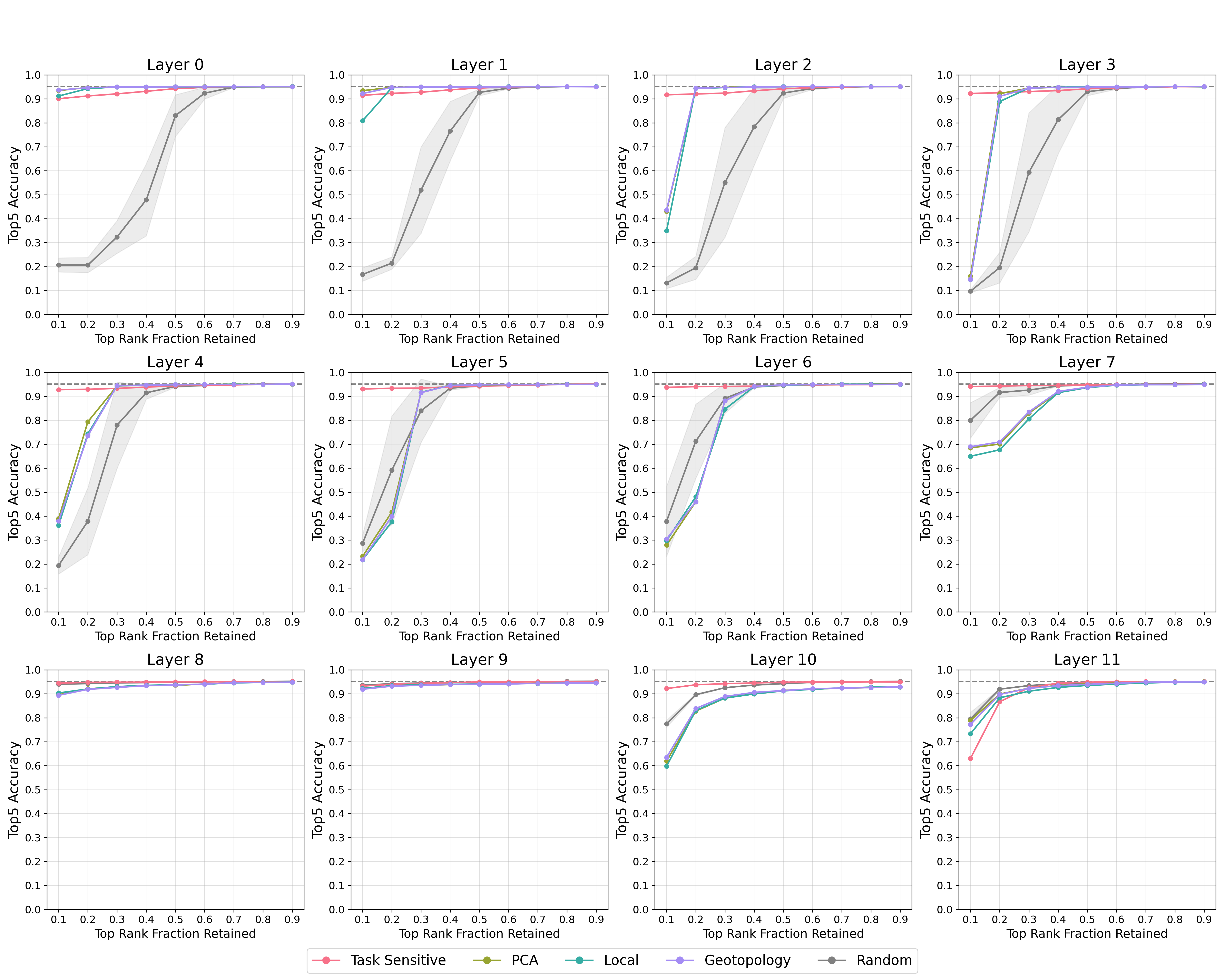}
    \caption{
        \textbf{Task-sensitive compression outperform geometric alternatives at most middle and late layers.}
        Per-layer top-5 ImageNet accuracy vs.\ retained rank fraction for five compression methods (ViT-B/16).
        At early layers, PCA and task-sensitivity perform similarly. They diverge from layers 4--6, task-sensitive projections preserve $>$93\% top-5 at 10\% retention while geometric alternatives fall below 35\%.
    }
    \label{fig:a2_utility_curves}
\end{figure}

The answer depends on depth (Fig~\ref{fig:a2_utility_curves}). At early layers (0--1), PCA substantially outperforms task-sensitive compression: at 10\% retention, PCA preserves 93.5\% top-5 accuracy at layer~0 versus 90.1\% for task-sensitivity. This likely suggests that early representations are dominated by a few high-variance principal components that also happen to be task-sensitive, so variance is a useful proxy for behavioral utility.
At middle layers (4--8), this relationship inverts dramatically.
At layer~6 with 10\% retention, task-sensitive compression preserves 93.8\% top-5 accuracy, while PCA preserves only 27.9\% . Local-neighborhood and geotopology-preserving projections perform comparably to PCA at these layers (29.8\% and 30.4\% top-5 at layer~6), confirming that the failure is not specific to PCA but reflects a general dissociation between geometric structure and functional sensitivity.
% At middle layers (4--8), this relationship inverts dramatically.
% At layer~6 with 10\% retention, task-sensitive compression preserves 76.4\% top-1 and 93.8\% top-5 accuracy, while PCA preserves only 16.2\% top-1 and 27.9\% top-5. Local-neighborhood and geotopology-preserving projections perform comparably to PCA at these layers (29.8\% and 30.4\% top-5 at layer~6), confirming that the failure is not specific to PCA but reflects a general dissociation between geometric structure and functional sensitivity.

This dissociation has a clear interpretation. In mid-later layers, the network's computation increasingly depends on directions that carry relatively little variance in the activation space - directions that PCA discards first. The task-sensitivity matrix identifies these directions because it is defined by the Jacobian of the downstream computation, not by the statistics of the representation itself.

The single-layer results show that compressibility varies across depth and some layers can be aggressively compressed while others cannot.
This motivates a global compression setting to investigate interactive and accumulative effects across layers. We allocate a fixed total rank budget across all layers (e.g., retain 30\% of the total dimensions summed over all layers). We compare a uniform allocation with a performance-aware allocation derived from the single-layer compression curves via dynamic programming, assigning more dimensions to layers where compression is costly and fewer to layers where it is cheap.

\begin{figure}[htbp]
    \centering
    \includegraphics[width=\linewidth]{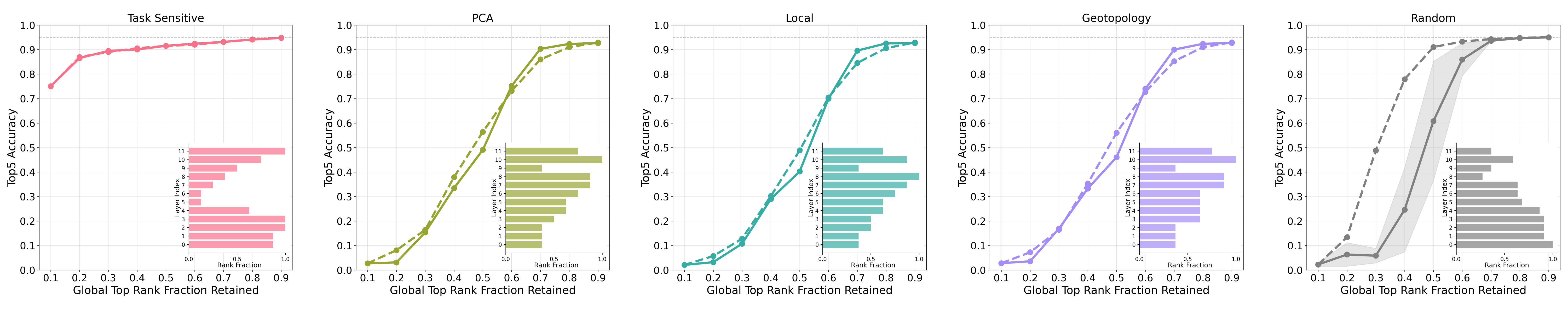}
    \caption{
        \textbf{Task-sensitive compression maintains a large advantage under global rank budgets.}
        Top-5 ImageNet accuracy when a total rank budget is distributed across layers via dynamic programming (solid) or uniform allocation (dashed).
        At 30\% global retention, task-sensitivity achieves 89.5\% top-5 vs.\ 15.3\% for PCA.
        Inset: optimal rank allocation across layers at $0.5$ retained fraction.
    }
    \label{fig:a3_global_curves_allocation}
\end{figure}

At 30\% global retention, task-sensitive compression achieves 89.5\% top-5 accuracy, compared to 15.3\% for PCA, 10.8\% for local geometry, and 16.9\% for geotopology (Figure \ref{fig:a3_global_curves_allocation}).
At 50\% global retention, the gap narrows but remains substantial: 91.6\% for task-sensitivity versus 49.1\% for PCA.
% At 30\% global retention, task-sensitive compression achieves 64.1\% top-1 and 89.5\% top-5 accuracy, compared to 4.9\%/15.3\% for PCA, 3.6\%/10.8\% for local geometry, and 6.0\%/16.9\% for geotopology (Figure \ref{fig:a3_global_curves_allocation}).
The geometric methods do not match task-sensitive compression until approximately 70--80\% of the total rank budget is retained, at which point all methods converge toward the uncompressed baseline. The dynamic allocation of retained rank across layers also indicates that task-sensitive compression is costly in early layers. These layers require more rank retention to maintain performance compared to the geometric methods.

Random projections also provide a useful comparison. At middle layers where task-sensitive compression dominates (layers 4--6), random projections perform poorly (20--38\% top-5 at 10\% retention), confirming that the task-sensitive directions are genuinely specific (Figure \ref{fig:a2_utility_curves}). At later layers (8–11), however, random projections recover surprisingly well (77.2\% and 94.1\% top-5 at 10\% retention), suggesting that task-sensitive information at these layers is more diffusely distributed across the representation and less concentrated in a small subspace.

\subsection{Task-sensitive compression alleviates adversarial vulnerability}
With evidence that task-sensitive compression can preserve performance, we next ask what other advantages compressed representations may offer. We start by investigating whether compressed representations yield advantages in terms of adversarial robustness. A leading hypothesis on the origins of such brittleness in networks is that it stems from their reliance on predictive but non-robust features in the data \citep{ilyas2019adversarial}. We ask how important these non-robust features are to task performance and whether our compression techniques alleviate adversarial vulnerability. Previous work has attempted to address this question with weight-space compression to mixed insights \citep{pavlitska2023relationship}. Our compression framework extends this hypothesis to the representational space: if networks encode directions that are sensitive to imperceptible image perturbations but not necessarily to the task, removing those directions should improve robustness. We evaluate robustness under the Projected Gradient Descent (PGD) attack \citep{madry2017towards} after compressing each layer's representations, measuring top-5 accuracy on the adversarial examples. More details on the implementation can be found in Appendix \ref{app:adv_expt}.

Figure \ref{fig:b_adversarial}a shows that compressing early-mid layers' representations improves robustness, especially under stronger compression. Notably, all compression methods provide robustness benefits, although task-sensitivity performs best at low ranks but its advantage diminishes over higher ranks. This suggests that structured compression in general is able to discard directions vulnerable to adversarial perturbations. Interestingly, we also note that the last layers of the network become more vulnerable to the attacks when compressed which suggests that task-sensitive and geometric information are deeply entangled with adversarially vulnerable features. This is consistent with the finding from \citet{fort2024ensemble} that attacks typically only fool later layers while earlier layers maintain their robustness. 

Finally, task-sensitive compression yields a substantial robustness advantage when applied across all layers of the network (Figure~\ref{fig:b_adversarial}b), one that the geometric compression methods do not provide. A caveat is that the adversarial examples were generated on the uncompressed network and then evaluated on the compressed ones. Part of the gain may therefore reflect reduced transferability of these attacks to the compressed networks rather than intrinsic robustness. Nonetheless, the comparison across methods is controlled: all compressed networks face the same fixed attacks at matched rank budgets, and only task-sensitive compression retains a consistent advantage. This suggests that directions with low task-sensitivity contribute to the vulnerability exposed by perturbations crafted on the full network, consistent with functional redundancy in the representation being a source of adversarial vulnerability.

\begin{figure}[htbp]
    \centering
    \begin{subfigure}[t]{0.76\linewidth}
        \centering
        \includegraphics[width=\linewidth]{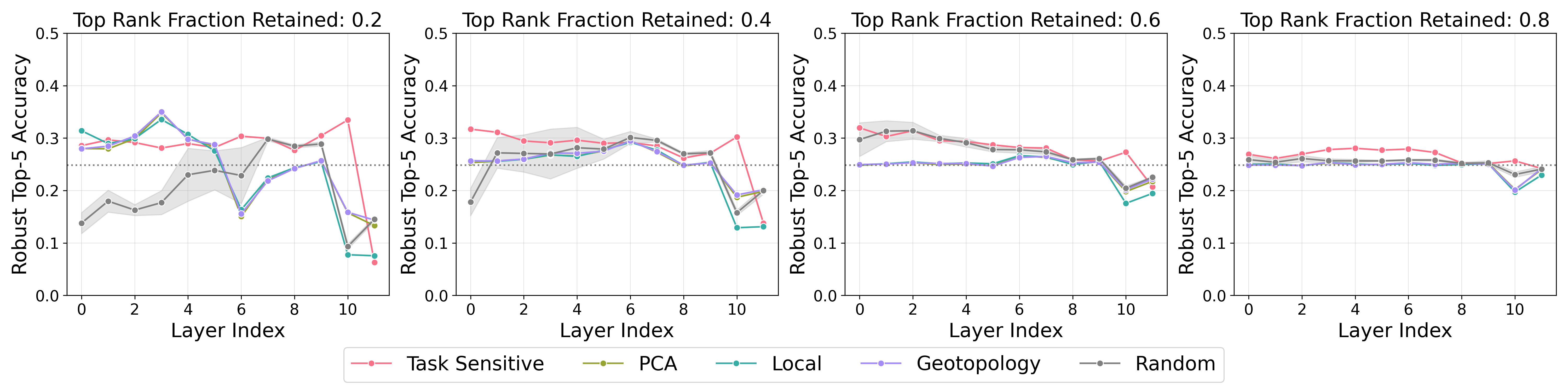}
        \caption{
            Layer-wise
        }
        \label{fig:b11_layerwise_adv}
    \end{subfigure}
    \hfill
    \begin{subfigure}[t]{0.21\linewidth}
        \centering
        
        \raisebox{0.2cm}
        {\includegraphics[width=\linewidth]{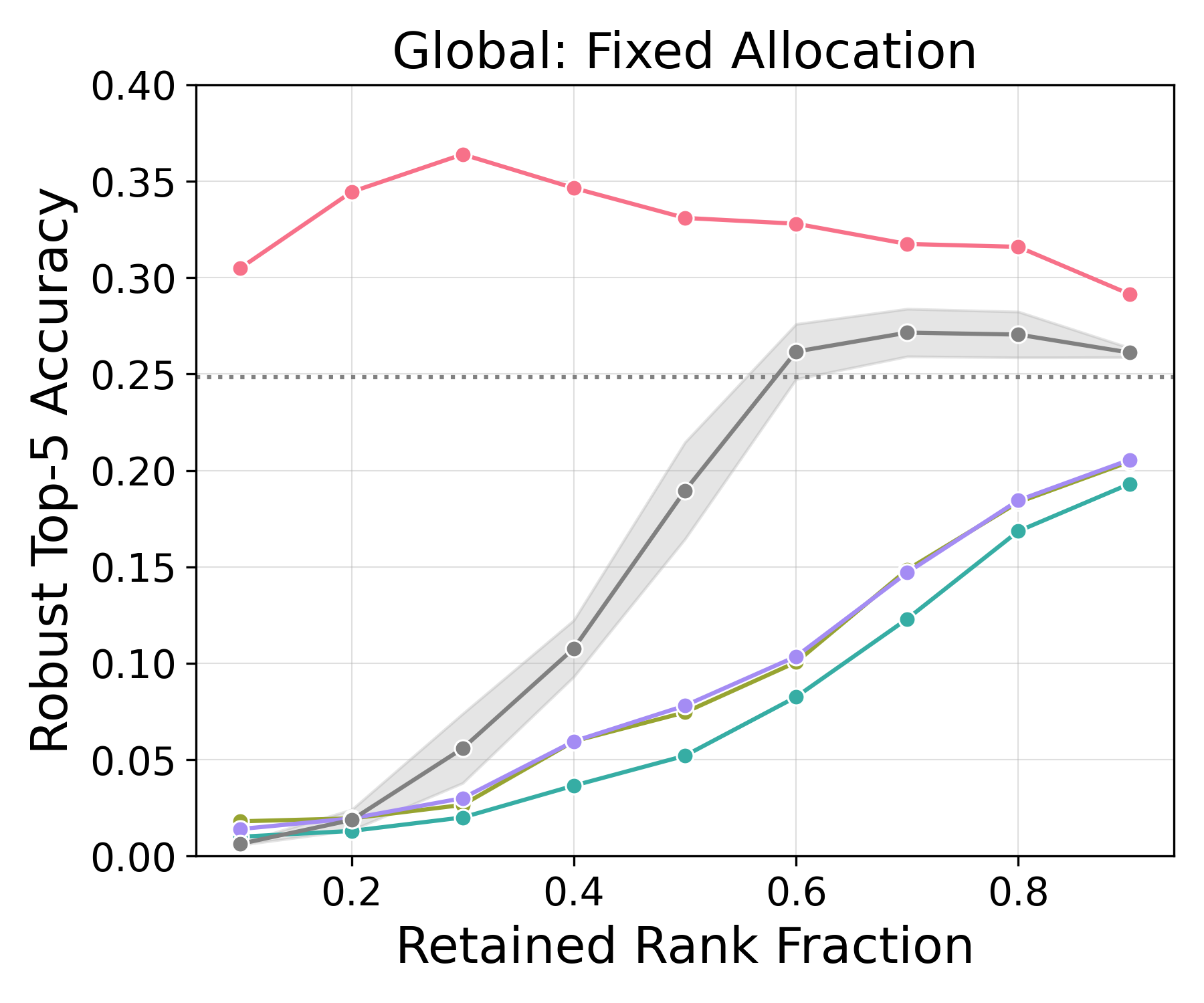}}
        \caption{
            Global
        }
        \label{fig:b12_global_adv}
    \end{subfigure}
    \caption{
        \textbf{Adversarial Vulnerability of Compressed Representations.}
        Adversarial robustness evaluated across projection methods and compression
        budgets under PGD-20 attack ($\epsilon = 4/255$, $\ell_\infty$).
        \subref{fig:b11_layerwise_adv}~Layer-wise view at four rank fractions.
        \subref{fig:b12_global_adv}~Global view under fixed allocation.
        Dashed line indicates uncompressed robust baseline.
    }
    \label{fig:b_adversarial}
\end{figure}
\subsection{Task-sensitive compression provides a natural basis for functionally-sensitive representational comparisons}
\label{subsec:rep-comparisons}

Standard representational similarity measures such as CKA \citep{kornblith2019similarity}, Procrustes distance \citep{williams2021generalized}, and RSA \citep{kriegeskorte2008representational} compare representations in their full ambient space. But if representations are compressible, i.e., if most of the ambient dimensionality is functionally inert as the preceding sections demonstrate, then full-space comparisons may be dominated by structure that plays no role in the network's computation. \citet{braun2025not} formalize this concern analytically in deep linear networks, representational similarity and functional similarity are dissociable, and full-space measures like CKA and RSA are often not driven by shared function.
\citet{cloos2024differentiable} demonstrate a complementary limitation empirically: by differentiating through similarity measures to maximize scores, they show that CKA can reach near-perfect values while failing to capture low-variance directions that carry task-sensitive information.
Together, these results suggest that full-space comparisons may conflate functionally sensitive alignment with alignment along behaviorally inert directions.

The task-sensitivity decomposition now provides a way to extract networks' functionally sensitive subspaces empirically and compare them instead of the full ambient space. We project each network's representation onto its own task-sensitive subspace (the top-$k$ eigenvectors of $F_\ell$) and onto its null subspace (the bottom-$k$ eigenvectors), then compute CKA separately in each. We note that these projections retain geometrically distinct structure (see Appendix~\ref{app:selfalignment}). If inter-network convergence is driven by shared functional structure, task-sensitive alignment should exceed null-space alignment.
If it is instead driven by shared inductive biases or generic statistical regularities unrelated to the task, the reverse should hold.

\paragraph{Task-sensitive alignment exceeds null-space alignment at later layers.}
We compute CKA between all pairs of four pretrained ViT networks (ViT-B/16, ViT-B/16 pretrained on ImageNet-21k, DeiT3-B/16 and ViT-L/16) at matched relative depths. At each layer, we extract the task-sensitivity matrix $F_\ell$ using the Jacobian of the pretrained classification head through all remaining layers, decompose into the top-$k$ and bottom-$k$ eigenvectors for a fixed $k$, project held-out test representations onto each subspace, and compute CKA separately (see Appendix~\ref{app:alignment} for more details). Figure~\ref{fig:alignment_combined}A--B shows the results averaged across all six network pairs for a range of subspace ranks $k \in \{25, 50, 100, 200, 300\}$. At early layers (panel A), task-sensitive and null-space alignment are comparable across all ranks, consistent with the observation that early representations are dominated by generic visual features that have not yet differentiated into task-specific structure. At later layers (panel B), a clear gap emerges: task-sensitive CKA consistently exceeds null-space CKA, and this gap is largest at low ranks where the decomposition isolates the most and least task-sensitive directions most sharply. This indicates that at later layers, networks are more aligned in their task-sensitive directions than in their task-irrelevant directions, an asymmetry that standard full-space CKA, which averages over all directions equally, cannot reveal.

\paragraph{Task-conditioned alignment reveals differential alignment for functionally distinct subspaces.}
The same framework extends beyond the binary task/null partition: replacing the classification head with a different task yields a different subspace and therefore a different alignment score over the same stimuli and networks. This enables \emph{task-conditioned} representational comparison, something standard similarity measures, which produce a single scalar per stimulus set, cannot do.
We demonstrate this using the cue-conflict stimuli of \citet{geirhos2018imagenet}, in which each image pairs one object's shape with another's texture (e.g., a cat silhouette rendered in elephant skin), providing independent shape and texture labels for the same set of images. This design is well suited for our purpose: because the two tasks share an identical stimulus set, their task-sensitivity matrices live in the same representational coordinate system and can be directly compared. For the same set of images and the same pair of networks, we compute two task-sensitivity matrices, one defined by a shape probe and one by a texture probe, and measure CKA separately in each subspace (see Appendix~\ref{app:alignment} for more details).
Figure~\ref{fig:alignment_combined}C shows the results averaged across all network pairs, with individual pairs connected by lines.
Texture-conditioned CKA substantially exceeds shape-conditioned CKA for every pair (texture: $0.51$--$0.89$; shape: $0.10$--$0.53$). This means that when two networks process the same images, they agree much more on directions that carry texture information than shape information. 
% A standard full-space CKA between these models would report a single value (typically $0.6$--$0.9$) that obscures this asymmetry entirely.

\begin{figure}[t]
\centering
\includegraphics[width=0.8\textwidth]{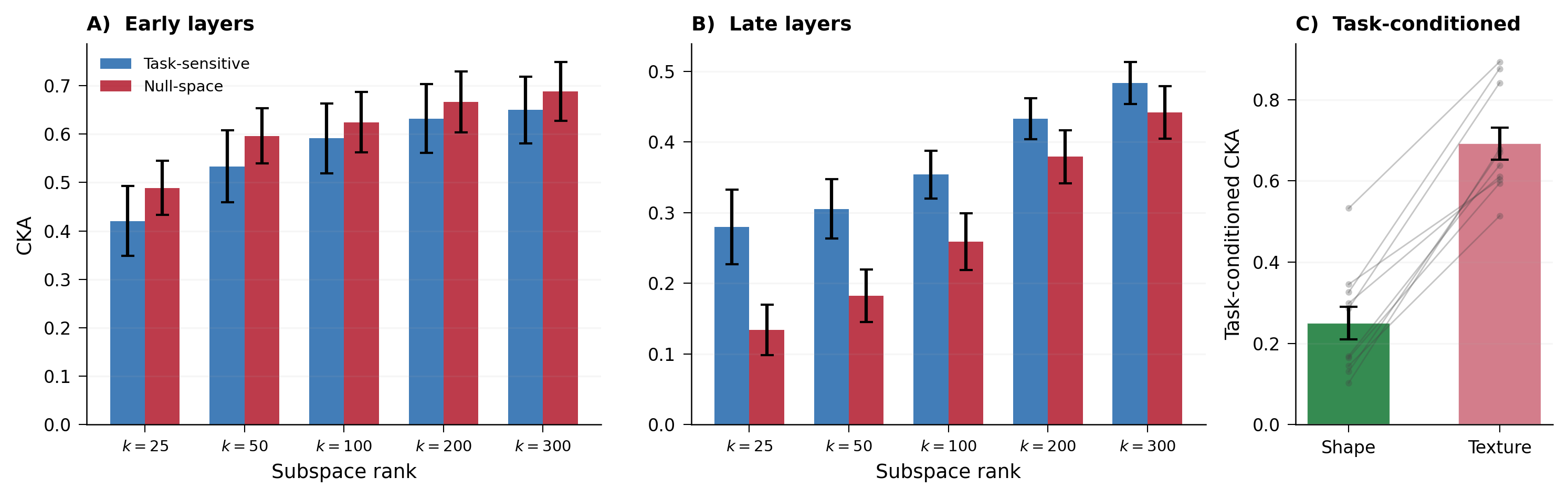}
\caption{\textbf{Task-conditioned representational comparison.} \textbf{A--B)} Task-sensitive CKA (blue) and null-space CKA (red) are comparable at early layers but diverge at later layers across all subspace ranks. \textbf{C)} Texture-conditioned CKA exceeds shape-conditioned CKA for every network pair.}
\label{fig:alignment_combined}
\end{figure}

\subsection{Task-sensitive decomposition recovers functionally distinct subspaces}
The task-sensitivity matrix $F_\ell$ identifies directions along which downstream computation is most sensitive, but is this decomposition meaningful: do task-sensitivity matrices defined by distinct tasks identify meaningfully distinct subspaces? We develop the idea of task-conditioned subspace alignment  from Section \ref{subsec:rep-comparisons} and assess how functionally distinct these identified subspaces are using the same cue-conflict stimuli.
\begin{wrapfigure}{r}{0.48\textwidth}
% \vspace{20pt}
\centering
\includegraphics[width=\linewidth]{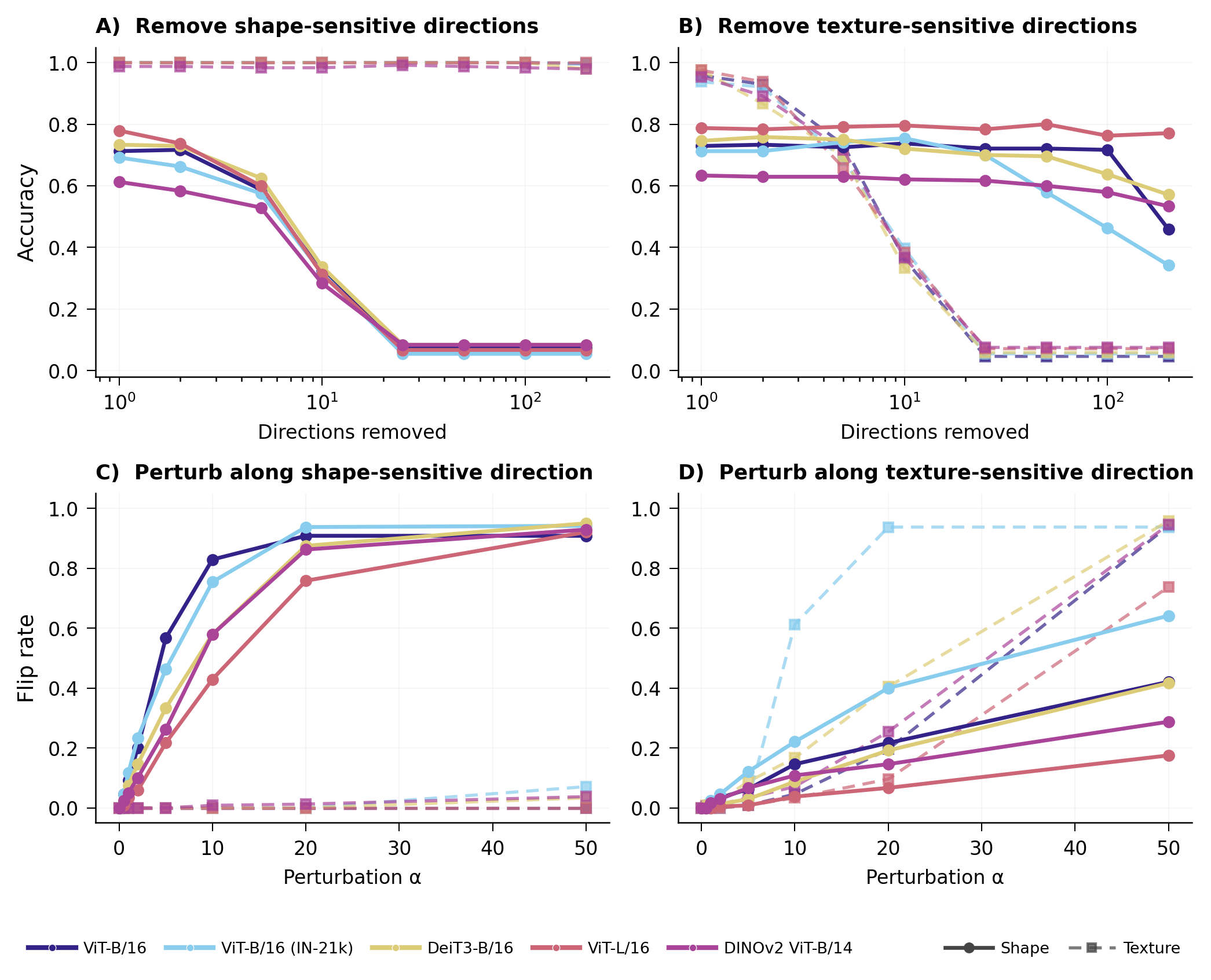}
\caption{\textbf{Task-sensitivity matrices identify functionally specific subspaces.} Solid: shape; dashed: texture. \textbf{A--B)} Ablating top eigenvectors of $F_{\mathrm{shape}}$ (A) or $F_{\mathrm{texture}}$ (B) selectively impairs the corresponding task. \textbf{C--D)} Perturbations along each task's top eigenvector flip that task's predictions first. }
\label{fig:double_dissociation}
\vspace{-20pt}
\end{wrapfigure}
% We test this using the cue-conflict stimuli of \citep{geirhos2018imagenet}, in which each image pairs one object's shape with another's texture (e.g., a cat silhouette rendered in elephant skin), providing independent shape and texture labels for the same set of images. This design is well suited for our purpose: because the two tasks share an identical stimulus set, their task-sensitivity matrices live in the same representational coordinate system and can be directly compared.

For each of five pretrained vision networks, we extract representations from the penultimate transformer block and train separate probes to predict shape and texture categories. We then compute the task-sensitivity matrix for each probe, yielding per-network shape-sensitive and texture-sensitive subspaces (see Appendix~\ref{app:shape_texture} for more details).
 
If the decomposition captures genuinely task-specific structure, then ablating shape-sensitive directions should selectively impair shape classification while leaving texture intact, and vice versa.
Figure~\ref{fig:double_dissociation} (panels A--B) confirms this across all five architectures: removing the top $k$ eigenvectors of $F_{\mathrm{shape}}$ causes shape accuracy to drop sharply while texture accuracy remains near ceiling, and removing the top eigenvectors of $F_{\mathrm{texture}}$ produces the complementary pattern.
As a further test, we perturb each test representation along the top eigenvector of $F_{\mathrm{shape}}$ or $F_{\mathrm{texture}}$ at increasing amplitudes and measure how frequently each probe's prediction changes (panels C--D).
Perturbations along the shape-sensitive direction cause shape predictions to flip at low perturbation levels, and perturbations along the texture-sensitive direction greatly flip texture predictions at low perturbations and some shape predictions at higher perturbations. 
Together, these results indicate that shape- and texture-sensitive subspaces are dissociable with low, but nonzero, overlap. 

\subsection{Task-sensitive subspaces as a basis for continual learning}
    The task-sensitive subspace identifies directions in representation space along which downstream computation is most sensitive. A natural application is continual learning: if the subspace that matters for a task can be identified, then future learning can be constrained to avoid interfering with it. Gradient Projection Memory (GPM; \citealt{saha2021gradient}) implements this idea using the SVD of the representation matrix to identify important subspaces for each task, then projecting gradient updates to lie orthogonal to them.
We propose Fisher GPM (FGPM), which replaces the SVD-based basis with the eigenvectors of the task-sensitivity matrix $F_\ell$.
Where GPM protects directions of high activation variance, FGPM protects directions of high downstream sensitivity.
These two criteria can diverge as we saw in earlier sections: a high-variance direction may carry no task-sensitive information, while a low-variance direction may be critical for downstream computation.

\begin{table}[t]
\centering
\caption{Continual learning with task-sensitive gradient projection (FGPM). We report average final accuracy (Acc, \%) and forgetting (Fgt, \%) on split CIFAR-10 (5 tasks) and split CIFAR-100 (20 tasks each). For GPM and FGPM, we select the best threshold $\varepsilon \in \{0.9, 0.95, 0.97\}$ per architecture. \textbf{Bold}: best non-oracle accuracy per row.}
\label{tab:continual_learning}
\resizebox{\textwidth}{!}{%
\small
\setlength{\tabcolsep}{3.5pt}
\begin{tabular}{l cccc ccc cccc ccc}
\toprule
& \multicolumn{7}{c}{CIFAR-10 (5 tasks)} & \multicolumn{7}{c}{CIFAR-100 (20 tasks)} \\
\cmidrule(lr){2-8} \cmidrule(lr){9-15}
& \multicolumn{4}{c}{Acc $\uparrow$} & \multicolumn{3}{c}{Fgt $\downarrow$} & \multicolumn{4}{c}{Acc $\uparrow$} & \multicolumn{3}{c}{Fgt $\downarrow$} \\
\cmidrule(lr){2-5} \cmidrule(lr){6-8} \cmidrule(lr){9-12} \cmidrule(lr){13-15}
& Naive & GPM & FGPM & Joint & Naive & GPM & FGPM & Naive & GPM & FGPM & Joint & Naive & GPM & FGPM \\
\midrule
AlexNet   & 73.5 & 86.4 & \textbf{86.6} & 94.2 & 24.5 & 7.6  & 8.3  & 44.9 & 61.3 & \textbf{65.3} & 84.2 & 39.3 & 15.7 & 15.1 \\
AlexNet-L & 61.3 & 71.5 & \textbf{78.8} & 90.5 & 34.8 & 24.3 & 11.8 & 54.6 & 52.0 & \textbf{60.5} & 79.4 & 18.8 & 11.1 & 5.7  \\
LeNet     & 83.6 & 83.9 & \textbf{86.9} & 92.1 & 10.5 & 8.8  & 6.2  & 53.8 & 66.6 & \textbf{70.0} & 80.9 & 27.5 & 8.2  & 8.0  \\
ResNet-18 & 76.7 & \textbf{89.2} & 87.8 & 94.3 & 21.8 & 1.7  & 7.6  & 45.2 & \textbf{65.6} & 64.3 & 85.7 & 38.0 & 11.0 & 12.4 \\
\bottomrule
\end{tabular}
}
\end{table}

We emphasize that our goal is not to benchmark against the full landscape of continual learning methods, which includes replay, regularization, and architecture-based approaches that differ from GPM in many dimensions simultaneously.
Rather, we use continual learning as a controlled test of a specific claim: that the task-sensitive subspace is a better basis for identifying what to protect than the variance-preserving subspace.
GPM is among the strongest methods in the subspace-projection family, performing competitively with state-of-the-art continual learning approaches across standard benchmarks \citep{saha2021gradient}, and provides the ideal comparison because the only difference between GPM and FGPM is the criterion used to select the protected subspace.
Everything else, including the gradient projection mechanism, the per-layer thresholding, and the sequential training protocol, is held constant.

 %For both GPM and FGPM, we select the best threshold $\varepsilon \in \{0.9, 0.95, 0.97\}$ 
We evaluate on split CIFAR-10 and split CIFAR-100 (5 \& 20 tasks each) across four architectures.
Table~\ref{tab:continual_learning} reports average final accuracy and forgetting for sequential fine-tuning (Naive), GPM, FGPM, and joint training (oracle upper bound).
(see Appendix~\ref{app:continual_learning} for more details). FGPM matches or exceeds GPM on six of eight architecture-dataset combinations.
The gains are largest on CIFAR-100, where the greater number of classes per task creates more pressure on representational capacity: FGPM improves over GPM by 4.0 points on AlexNet, 8.5 on AlexNet-L, and 3.4 on LeNet.
On CIFAR-10, FGPM outperforms GPM on three of four architectures, with AlexNet-L showing the largest gain (+7.3 points).
The exception in both datasets is ResNet-18, where GPM retains an advantage.
We hypothesize that this reflects an interaction between batch normalization and the Fisher estimate: the per-sample Jacobian does not account for the running statistics maintained by batch normalization layers, and this approximation error may compound through the deeper residual architecture.
 
These results demonstrate that the notion of compressibility developed in the preceding sections transfers directly to continual learning.
Protecting the behaviorally potent subspace rather than the high-variance subspace yields better preservation of prior-task performance, consistent with the finding that task-sensitive and variance-preserving projections identify increasingly different subspaces in later layers.

\section{Discussion}
Our central finding is that deep network representations are dramatically compressible: at most layers, projecting onto a small fraction of the task-sensitive subspace preserves nearly all downstream performance. 

These results connect to several lines of work on compression and representation geometry. In weight space, the contrast between geometric and performance-aware criteria has a long history. Magnitude-based pruning ranks parameters by a property of the weights themselves \citep{han2015deep}, and low-rank factorization via SVD compresses weight matrices by preserving their dominant singular directions \citep{denton2014exploiting}. Other methods instead estimate a parameter's effect on the task directly \citep{lecun1989optimal, hassibi1993optimal}. Recent work in Large Language Models (LLMs) has also revived the distinction between geometry-preserving and task-sensitive compression \citep{sharma2024truth, hsu2022language}. The overall picture in weight space is mixed, however: comparative studies have not identified a clear winner among selection criteria \citep{galeStateSparsityDeep2019}. In representation space, the picture is sharper: task-sensitive projections yield performance advantages over PCA, local-neighborhood and geotopology projections. 
Our layer-wise profile also fits the tunnel effect studied in \citet{masarczyk2023tunnel}, in which the later layers of sufficiently deep networks compress the representations built by earlier layers. The very low dimensionality of the task-sensitive subspace is in line with this picture. Our observation that variance-based compression can struggle to retain task performance has parallels in LLM literature too \citep{li2026tracing}. 
Together, these results show the functional advantage of task-sensitive compression over geometric ones for mid-late layers in deep networks. Further, the task-sensitive compressibility we observe has direct consequences for improving adversarial robustness, identifying functionally distinct spines within the same representation, grounding representational comparisons in behaviorally relevant structure, and for protecting prior knowledge during continual learning.

\paragraph{Limitations} The task-sensitivity matrix is a second-order approximation that captures the linear sensitivity of downstream computation to representational perturbations; it does not account for nonlinear interactions across layers, and its estimation quality may degrade for layers far from the output. Additionally, subspace dimensionality and estimation quality may vary across tasks. Differences in network capacity and architecture may confound comparisons of compressibility across networks, motivating more controlled cross-architecture analyses.

\paragraph{Future Work}Looking forward, the framework naturally extends to settings where multiple tasks share a single representation. LLMs could be investigated to determine whether different capabilities (e.g. factual recall, syntactic processing, reasoning) occupy distinct or overlapping functional spines, and whether the degree of overlap predicts interference during fine-tuning. Extending the approach to biological neural recordings, where the downstream computation is unknown and must itself be estimated, would connect this work to longstanding questions about neural coding efficiency and the functional interpretation of population activity. More broadly, the observation that representations are dramatically compressible raises questions pertaining to theories of representation learning: what determines the size of the task-sensitive subspace, how do these functional spines evolve during training, and does the remaining capacity serve a purpose (e.g. plasticity, future task transfer) that our current framework does not capture?

\begin{ack}
This work was supported by a Coefficient Giving (formerly Open Philanthropy) Grant to M.K. The authors declare no competing interests.
\end{ack}
\newpage
\bibliographystyle{plainnat}
\bibliography{references}

\newpage 
\appendix 
\section{Appendix}
\counterwithin{figure}{section}
\counterwithin{table}{section}
\renewcommand{\thefigure}{\thesection.\arabic{figure}}
\renewcommand{\thetable}{\thesection.\arabic{table}}
\subsection{Compression Technical Details}
\label{app:more_methods}

\subsubsection{PCA Projection}

The notion of compressing representations through PCA is also closely related to preserving global Euclidean geometry. If representational geometry is summarized by the squared Euclidean dissimilarity matrix $D^{(2)}_{ij} = \|z_i - z_j\|_2^2$, then classical multidimensional scaling on $D^{(2)}$ recovers exactly the PCA coordinates of the centered data.
Thus, under Euclidean distances, variance-preserving compression can equivalently be interpreted as preserving the dominant global shape of the representational geometry.
PCA therefore provides a natural geometry-preserving baseline against which to compare task-sensitivity--based compression.

\subsubsection{Geotopological Projection}
Let $d_{ij} = \|z_i - z_j\|_2$ denote pairwise representational distances.
We apply the piecewise-linear geotopological (GT) transform with lower and upper thresholds $l < u$ (set as quantiles of the empirical distance distribution):
$\mathrm{GT}_{l,u}(d) = 0$ if $d \le l$,
$\mathrm{GT}_{l,u}(d) = (d - l)/(u - l)$ if $l < d < u$, and
$\mathrm{GT}_{l,u}(d) = 1$ if $d \ge u$.
The transformed values define a weighted graph with edge weights $W^{\mathrm{GT}}_{ij} = \mathrm{GT}_{l,u}(d_{ij})$.
Letting $L^{\mathrm{GT}} = D^{\mathrm{GT}} - W^{\mathrm{GT}}$ denote the graph Laplacian with degree matrix $D^{\mathrm{GT}}_{ii} = \sum_j W^{\mathrm{GT}}_{ij}$, the geotopology matrix is $M_\ell^{\mathrm{GT}} = 2\, Z_\ell^\top L^{\mathrm{GT}} Z_\ell$.
For any direction $v$, the quadratic form $v^\top M_\ell^{\mathrm{GT}}\, v = \sum_{i,j} W^{\mathrm{GT}}_{ij}(v^\top(z_i - z_j))^2$ measures how strongly that direction preserves the transformed geotopological relations.
The top eigenvectors of $M_\ell^{\mathrm{GT}}$ define the subspace that best preserves this structure.
If $M_\ell^{\mathrm{GT}} = R \Omega R^\top$, the rank-$r$ geotopology-preserving projector is $\Pi_r^{\mathrm{GT}} = R_r R_r^\top$, where $R_r$ contains the top $r$ eigenvectors. In our implementation, we use the values $l=0.2$ and $u=0.8$.

\subsection{Evaluating Compressed Representations}
\label{app:accuracy_eval}
We evaluate task performance under ImageNet classification.

\paragraph{Layer-wise evaluation.} For each transformer layer $\ell \in \{0, \ldots, 11\}$
independently, we project the hidden state $z_\ell$ onto the top-$r$ subspace defined by
each compression criterion while leaving all other layers unmodified. We sweep the retained
rank fraction $r/d \in \{0.1, 0.2, \ldots, 0.9\}$ where $d = 768$ is the hidden dimension
of ViT-B/16. After projection, the representation is renormalized to match the Frobenius
norm of the original activation before being passed to subsequent layers. This isolates the
per-layer contribution to task performance and allows us to identify which layers are most
and least compressible under each criterion. Results are reported as the top-5 accuracy
difference relative to the uncompressed baseline $(\Delta\text{top-5})$, so that zero
indicates no degradation and positive values indicate compression marginally exceeds
baseline performance.

\paragraph{Global evaluation.} All transformer layers are compressed simultaneously under a
fixed total rank budget $r_{\text{total}} = \lfloor \rho \cdot d \cdot L \rfloor$, where
$\rho \in \{0.1, 0.2, \ldots, 0.9\}$ is the global retained fraction and $L = 12$ is the
number of transformer layers. We consider two rank allocation strategies across layers:

\begin{itemize}
    \item \textbf{Fixed allocation:} Each layer independently retains the same fraction
    $\rho$ of its $d$ dimensions, i.e., $k_\ell = \lfloor \rho \cdot d \rfloor$ for all
    $\ell$.

    \item \textbf{Utility-curve allocation:} Rank is allocated across layers via a
    dynamic programming procedure that maximizes total utility under the global budget
    constraint. The utility of retaining rank $r$ at layer $\ell$ is defined as the
    top-5 accuracy of the single-layer compression at that rank, estimated from the
    layer-wise evaluation above. This allows layers that are more compressible to receive
    fewer dimensions, concentrating capacity in layers where compression is most costly.
\end{itemize}

Global top-5 accuracy is reported as a function of $\rho$ for each compression method
under both allocation strategies, with the uncompressed baseline shown as a dashed line
(Figure~\ref{fig:a3_global_curves_allocation} in the main text). 

All layerwise and global evaluations use the same
class-balanced subset of 10,000 ImageNet validation images that is separate from the class-balanced 2,000 ImageNet images used to build the projector matrices.

\subsection{Task-sensitive compressibility across architectures}
\label{app:arch_generalization}

\begin{figure}[htbp]
    \centering
    \includegraphics[width=\linewidth]{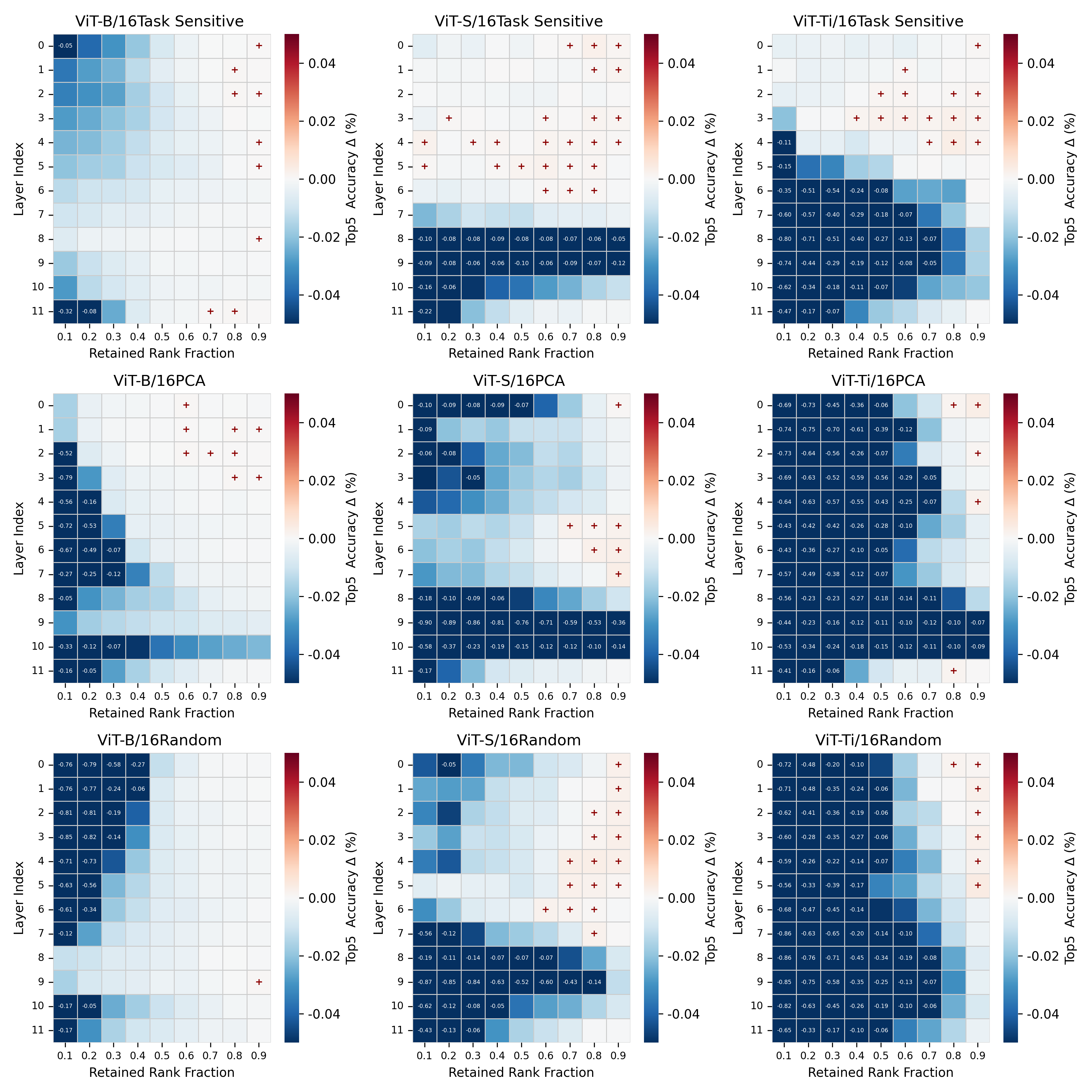}
    \caption{
        \textbf{Layer-wise compressibility across ViT architectures.}
        Top-5 ImageNet accuracy difference between compressed and full model as a function
        of layer index and retained rank fraction, shown for ViT-B/16, ViT-S/16, and
        ViT-Ti/16 (columns) under task-sensitive, PCA, and random projections (rows).
        Cells marked \texttt{+} indicate accuracy exceeding the uncompressed baseline.
        The compressibility advantage of task-sensitivity over geometric alternatives
        generalizes across all three architectures.
    }
    \label{fig:sa1_many_models}
\end{figure}

Figure~\ref{fig:sa1_many_models} extends the layer-wise compression analysis from Section~2.1
to two additional ViT variants: ViT-S/16 and ViT-Ti/16. Results are shown for task-sensitivity,
PCA, and random projections across all three architectures.

The pattern observed for ViT-B/16 in the main text generalizes broadly. Under task-sensitivity
projection, most layers preserve near-baseline top-5 accuracy across a wide range of retained
rank fractions, with accuracy losses concentrated at very low ranks in early and final layers.
Cells marked \texttt{+} indicate cases where compression marginally exceeds the uncompressed
baseline, consistent with the hypothesis that projecting out low-sensitivity directions removes
variation that slightly interferes with downstream computation. PCA projections show more
variable performance across layers and architectures, with larger accuracy drops at
intermediate ranks, particularly in ViT-S/16 and ViT-Ti/16 where the gap between
variance-preserving and task-sensitive compression is more pronounced. Random projections
degrade performance steeply at low ranks across all architectures and layers, confirming that
the compressibility observed under task-sensitivity is not a trivial consequence of
dimensionality reduction per se.

\subsection{Adversarial Vulnerability experiments}
\label{app:adv_expt}

\subsubsection{Setup}

We evaluate adversarial robustness by applying single-layer and multi-layer representation
projections to a pretrained ViT-B/16 and measuring classification accuracy on adversarially
perturbed inputs. Projections are applied mid-network: the hidden state at a given layer
$z_\ell \in \mathbb{R}^d$ is projected onto the retained subspace via $\tilde{z}_\ell =
\Pi_r z_\ell$, where $\Pi_r = U_r U_r^\top$ is the rank-$r$ projector defined by each
compression criterion. After projection, the representation is renormalized to match the
Frobenius norm of the original activation, following the same convention used throughout the
clean accuracy evaluations. All subsequent layers receive the compressed representation
unmodified.

Adversarial evaluation uses the same class-balanced subset of 2,000 ImageNet validation images
as the clean accuracy experiments, drawn with a fixed random seed of $0$. A key implementation detail here is that the adversarial examples were generated on the base model before any projections were applied. This was a more sound option as our motivation for the experiment is to uncover the representational components of adversarial vulnerability and not to present a novel robust method.

\subsubsection{Attack Construction}

We use Projected Gradient Descent \citep{madry2017towards} as the primary attack. Attacks are
generated with perturbation budget $\epsilon = 4/255$ in the $\ell_\infty$ norm, step size
$\alpha = 1/255$, and 20 iterations (PGD-20). Attacks are crafted against the
\emph{uncompressed} full model and cached prior to evaluation. The same set of adversarial
images is then used to evaluate all compression methods and rank fractions. 

We report robust top-5 accuracy as the primary metric. The uncompressed model's robust
top-5 accuracy under the same attack is shown as a dotted reference line in all figures
(Figure~\ref{fig:b_adversarial}).

\subsubsection{Robustness Evaluation: Layer-wise and Global}

We evaluate robustness under the same two modes used for clean accuracy (Figure \ref{fig:b_adversarial} in the
main text):

\paragraph{Layer-wise evaluation.} For each transformer layer $\ell \in \{0, \ldots, 11\}$
independently, we project the hidden state $z_\ell$ onto the top-$r$ subspace defined by
each compression criterion while leaving all other layers unmodified. We sweep the retained
rank fraction $r/d \in \{0.1, 0.2, \ldots, 0.9\}$ where $d = 768$ is the hidden dimension
of ViT-B/16. This isolates the per-layer contribution to adversarial vulnerability and
allows us to identify which layers are most affected by structured compression.
Figure~\ref{fig:b_adversarial}a shows robust top-5 accuracy as a function of layer index
for four fixed rank fractions.

\paragraph{Global evaluation.} All transformer layers are compressed simultaneously under a
fixed total rank budget $r_{\text{total}}$.  We
evaluate a fixed allocation baseline in which each layer independently retains the same
fraction of its dimensions. Figure~\ref{fig:b_adversarial}b reports robust top-5
accuracy as a function of total retained rank fraction for each compression method with the uncompressed robust baseline shown as a dashed line.

\subsection{Shape--texture dissociation experiments}
\label{app:shape_texture}

\subsubsection{Stimuli}

We use the cue-conflict stimuli from \citet{geirhos2018imagenet}, in which each image is generated by applying neural style transfer to combine the shape of one object category with the texture of another.
The stimulus set spans 16 categories (airplane, bear, bicycle, bird, boat, bottle, car, cat, chair, clock, dog, elephant, keyboard, knife, oven, truck).
We retain only images where shape and texture categories differ, resulting in 1,200 cue-conflict images.
Each image carries two independent labels: a shape label (the category whose silhouette is preserved) and a texture label (the category whose surface statistics are transferred).
Images are preprocessed with standard ImageNet normalization (resize to 256, center crop to 224, channel-wise mean and standard deviation).

\subsubsection{Models}

We evaluate five pretrained vision transformer models, all frozen throughout the experiment:
ViT-B/16 trained on ImageNet-1k (\texttt{vit\_base\_patch16\_224}, timm);
ViT-B/16 pretrained on ImageNet-21k and fine-tuned on ImageNet-1k (\texttt{vit\_base\_patch16\_224.augreg\_in21k\_ft\_in1k}, timm);
DeiT3-B/16 trained with the 3-Augment recipe (\texttt{deit3\_base\_patch16\_224}, timm);
ViT-L/16 trained on ImageNet-1k (\texttt{vit\_large\_patch16\_224}, timm), which has 24 blocks and hidden dimension 1024 compared to 12 blocks and dimension 768 for the Base models;
and DINOv2 ViT-B/14 trained with self-supervised learning (\texttt{dinov2\_vitb14}, torch.hub).
Features are extracted at the penultimate transformer block (block 10 for 12-block models, block 22 for ViT-L/16) as the CLS token output.

\subsubsection{Probes and task-sensitivity matrices}

For each model, we train two separate probes on the frozen CLS token features: one predicting shape category and one predicting texture category (16 classes each).
Probes are trained with AdamW (learning rate $5 \times 10^{-3}$, weight decay $10^{-4}$) and cosine annealing for 200 epochs on 80\% of the images (960 samples), with the remaining 20\% (240 samples) held out for evaluation.
The task-sensitivity matrix for each probe is computed on the training split as $F = \frac{1}{N} \sum_{i=1}^{N} \nabla_{z} \mathcal{L}(z_i, y_i) \, \nabla_{z} \mathcal{L}(z_i, y_i)^\top$, where $\mathcal{L}$ is the cross-entropy loss and $z_i$ is the CLS token at the penultimate layer.
Gradients are computed via backpropagation through the probe only, and the matrix is symmetrized as $F \leftarrow \frac{1}{2}(F + F^\top)$.

\subsubsection{Ablation and perturbation protocols}

For ablation, we remove the top $k \in \{1, 2, 5, 10, 25, 50, 100, 200\}$ eigenvectors of $F_{\mathrm{shape}}$ (or $F_{\mathrm{texture}}$) by projecting test representations onto the orthogonal complement: $\tilde{z} = (I - V_{:,1:k} V_{:,1:k}^\top) z$.
Both probes are then evaluated on the modified representations.
For perturbation, we perturb each test representation along the top eigenvector $v_1$ of $F_{\mathrm{shape}}$ (or $F_{\mathrm{texture}}$) at amplitudes $\alpha \in \{0, 0.5, 1, 2, 5, 10, 20, 50\}$: $\tilde{z} = z + \alpha \cdot \frac{\bar{\|z\|}}{100} \cdot v_1$, where $\bar{\|z\|}$ is the mean feature norm across the test set, and we measure the fraction of images for which each probe's prediction changes.
All results are reported on the held-out test split.

\subsubsection{Per-model results}

Table~\ref{tab:shape_texture_probes} reports probe accuracy and task-sensitivity subspace statistics.
Texture probe accuracy is near-perfect ($\geq$98\%) across all models, while shape probe accuracy ranges from 62\% to 79\%, consistent with the texture bias of supervised ImageNet models \citep{geirhos2018imagenet}.
The effective rank of $F_{\mathrm{shape}}$ is consistently between 5.7 and 8.4, indicating that shape classification depends on a small but non-trivial number of directions.
The effective rank of $F_{\mathrm{texture}}$ varies similarly between 5.4 to 7.7. We also compute the principal angles between the shape-sensitive and texture sensitive subspaces and find that the median angle remains between $82.7^\circ$ and $84.0^\circ$ for every model suggesting substantially distinct subspaces.

\begin{table}[t]
\centering
\caption{Shape and texture space statistics}
\label{tab:shape_texture_probes}
\small
\setlength{\tabcolsep}{4pt}
\begin{tabular}{l cc cc cc cc}
\toprule
& \multicolumn{2}{c}{Shape} 
& \multicolumn{2}{c}{Texture} 
& \multicolumn{2}{c}{Eff.\ rank (PR)} 
& \multicolumn{2}{c}{Principal angle ($^\circ$)} \\
\cmidrule(lr){2-3} \cmidrule(lr){4-5} 
\cmidrule(lr){6-7} \cmidrule(lr){8-9}
Model & Train & Test & Train & Test 
& $F_{\mathrm{shape}}$ & $F_{\mathrm{texture}}$
& Median & Range \\
\midrule
ViT-B/16          & 100.0 & 72.5 & 100.0 & 100.0 & 6.9 & 7.6 & 83.2 & 77.8--89.7 \\
ViT-B/16 (IN-21k) & 100.0 & 71.2 & 100.0 & 100.0 & 8.4 & 7.7 & 82.7 & 72.7--88.7 \\
DeiT3-B/16        & 100.0 & 75.0 & 100.0 & 100.0 & 6.9 & 5.4 & 83.3 & 80.0--89.9 \\
ViT-L/16          & 100.0 & 79.2 & 100.0 & 100.0 & 5.7 & 5.4 & 84.0 & 78.4--89.8 \\
DINOv2 ViT-B/14   & 100.0 & 62.1 & 100.0 & 98.3  & 7.1 & 6.1 & 83.4 & 73.8--89.7 \\
\bottomrule
\end{tabular}

\vspace{4pt}
{\footnotesize Accuracy in \%. Train accuracy is from the final epoch; test accuracy is on the held-out 20\%.  Effective rank is the participation ratio of the eigenvalue spectrum of $F$. Principal angles between the top-10 shape- and texture-sensitive subspaces are reported in degrees.}
\end{table}

\subsection{Continual learning experiments}
\label{app:continual_learning}

\subsubsection{Task construction}

We evaluate on two standard continual learning benchmarks derived from CIFAR \citep{krizhevsky2009learning}. Split CIFAR-10 partitions the 10 classes into 5 sequential tasks of 2 classes each. Split CIFAR-100 partitions the 100 classes into 20 sequential tasks of 5 classes each. In both cases, classes are assigned to tasks in their original label order (i.e., classes 0--1 form task 1, classes 2--3 form task 2, etc.) and no data augmentation is applied beyond normalization to the channel-wise CIFAR mean and standard deviation. All experiments use a single fixed random seed.

The setup is task-incremental: each task has its own classification head, and the task identity is provided at test time. This isolates the forgetting problem from the task-identification problem and is standard in the GPM literature \citep{saha2021gradient}.

\subsubsection{Architectures}

We evaluate four architectures, each with batch normalization applied after every convolutional and fully connected layer (excluding the final classification heads). All architectures are adapted for $32 \times 32$ input resolution.

\paragraph{AlexNet.} Three convolutional layers ($3 \to 32 \to 64 \to 128$ channels, $3 \times 3$ kernels, padding 1) each followed by batch normalization, ReLU, and $2 \times 2$ max-pooling, then two fully connected layers ($2048 \to 256 \to 256$) each followed by batch normalization, ReLU, and dropout ($p = 0.25$). Protected layers: \texttt{conv1}, \texttt{conv2}, \texttt{conv3}, \texttt{fc1}, \texttt{fc2}.

\paragraph{AlexNet-L.} A wider variant matching the architecture used in the original GPM paper \citep{saha2021gradient}. Three convolutional layers ($3 \to 64$, kernel $4 \times 4$; $64 \to 128$, kernel $3 \times 3$; $128 \to 256$, kernel $2 \times 2$) each followed by batch normalization, ReLU, dropout ($p = 0.2$), and $2 \times 2$ max-pooling, then two fully connected layers (both 2048-dimensional) each followed by batch normalization, ReLU, and dropout ($p = 0.5$). Protected layers: \texttt{conv1}, \texttt{conv2}, \texttt{conv3}, \texttt{fc1}, \texttt{fc2}.

\paragraph{LeNet.} Two convolutional layers ($3 \to 20 \to 50$ channels, $5 \times 5$ kernels) each followed by batch normalization, ReLU, and $2 \times 2$ max-pooling, then two fully connected layers ($1250 \to 800 \to 500$) each followed by batch normalization, ReLU, and dropout ($p = 0.25$). Protected layers: \texttt{conv1}, \texttt{conv2}, \texttt{fc1}, \texttt{fc2}.

\paragraph{ResNet-18.} A standard ResNet-18 \citep{he2016deep} adapted for $32 \times 32$ inputs: the initial $7 \times 7$ convolution is replaced with a $3 \times 3$ convolution and the initial max-pooling layer is removed. Block configuration is $[2, 2, 2, 2]$ with channels $[64, 128, 256, 512]$. A fully connected layer ($512 \to 512$) with batch normalization and ReLU is inserted before the task-specific heads. All convolutional layers in the residual blocks (16 total), the initial convolution, and \texttt{fc1} are protected.

\subsubsection{Training protocol}

All models are trained with SGD (learning rate 0.01, momentum 0.9, no weight decay) for 50 epochs per task with a batch size of 64. Tasks are presented sequentially; after training on each task, the subspace basis is computed and accumulated before proceeding to the next task. The joint training baseline uses the same optimizer and trains for $50 \times T$ total epochs (where $T$ is the number of tasks), cycling through all tasks at each gradient step.

\subsubsection{Subspace basis construction}

\paragraph{GPM.} Following \citet{saha2021gradient}, the GPM basis for each protected layer is constructed from the input activations collected during a forward pass over the training data of the current task. For convolutional layers, inputs are unfolded into patch matrices. The covariance matrix of the collected activations is computed, and its eigenvectors are extracted via eigendecomposition. The number of retained eigenvectors is determined by the energy threshold $\varepsilon$: the smallest $r$ such that the top $r$ eigenvalues account for at least a fraction $\varepsilon$ of the total variance. After each task, the new basis vectors are merged with the existing basis via QR orthogonalization, ensuring that the accumulated basis spans the union of all per-task subspaces.

\paragraph{FGPM.} The FGPM basis replaces input activations with a Jacobian-weighted quantity. For each protected layer, the gradients of the loss with respect to the layer's output activations are captured via backward hooks during a forward-backward pass over the current task's training data. Let $G \in \mathbb{R}^{N \times c_{\mathrm{out}}}$ denote the matrix of collected output gradients (row-normalized to unit norm) and let $W \in \mathbb{R}^{c_{\mathrm{out}} \times c_{\mathrm{in}}}$ denote the layer's weight matrix (reshaped for convolutional layers). The FGPM computes $H = G W / \sqrt{N}$ and extracts its right singular vectors via SVD. These right singular vectors lie in the input space of the layer and identify the directions along which perturbations most affect the downstream loss, approximating the top eigenvectors of the task-sensitivity matrix $F_\ell$ defined in the main text. The energy threshold and basis accumulation procedure are identical to GPM.

\paragraph{Threshold selection.} We report results for the best threshold $\varepsilon \in \{0.9, 0.95, 0.97\}$ that was selected separately for each method (GPM and FGPM), architecture and dataset. The presented results are each condition's performance on a held-out validation set. In all cases, the threshold is incremented by $0.003$ after each task to allow the protected subspace to grow slightly as more tasks are learned, following the protocol in \citet{saha2021gradient}. Up to 1500 separate samples from the current task's training set are used for basis construction.

\subsubsection{Evaluation metrics}

Let $a_{t,j}$ denote the test accuracy on task $j$ after training on task $t$, and let $T$ denote the total number of tasks.

\paragraph{Average final accuracy.} The mean accuracy across all tasks after training on the final task: $\mathrm{Acc} = \frac{1}{T} \sum_{j=1}^{T} a_{T,j}$.

\paragraph{Forgetting.} The average drop from peak accuracy to final accuracy across all tasks except the last:
\[
\mathrm{Fgt} = \frac{1}{T-1} \sum_{j=1}^{T-1} \Big( \max_{t \in \{1,\dots,T\}} a_{t,j} - a_{T,j} \Big).
\]

\subsubsection{Seed variability of GPM and FGPM}
In order to evaluate how robust FGPM's performance advantage is over GPM, we perform a multi-seed comparison at the same fixed base threshold $\epsilon=0.97$, while retaining the original $+0.003$-per-task schedule. The entries in Table \ref{tab:gpm_fgpm_multiseed} are final test accuracies reported as mean $\pm$ sample standard deviation over three paired seeds. At this fixed operating point, FGPM improves mean final accuracy in seven of eight settings, with particularly consistent gains for AlexNet-L and LeNet. The additional seeds also show that the CIFAR-10 ResNet-18 exception is reproducible and the CIFAR-100 AlexNet difference is small relative to its variability. 

\begin{table}[t]
\centering
\caption{Multi-seed comparison of GPM and FGPM across architectures and datasets.}
\label{tab:gpm_fgpm_multiseed}
\begin{tabular}{llccc}
\toprule
Dataset & Architecture & GPM & FGPM & FGPM $-$ GPM \\
\midrule
CIFAR-10  & AlexNet   & 72.65 $\pm$ 3.47 & 86.53 $\pm$ 0.29 & +13.88 $\pm$ 3.35 \\
CIFAR-10  & AlexNet-L & 69.56 $\pm$ 0.66 & 76.00 $\pm$ 0.17 & +6.44 $\pm$ 0.81  \\
CIFAR-10  & LeNet     & 77.74 $\pm$ 1.21 & 86.72 $\pm$ 0.22 & +8.98 $\pm$ 1.10  \\
CIFAR-10  & ResNet-18 & 89.17 $\pm$ 0.44 & 87.16 $\pm$ 0.57 & $-$2.01 $\pm$ 0.59 \\
CIFAR-100 & AlexNet   & 62.18 $\pm$ 1.48 & 64.46 $\pm$ 1.71 & +2.27 $\pm$ 2.91  \\
CIFAR-100 & AlexNet-L & 44.96 $\pm$ 2.28 & 53.66 $\pm$ 1.82 & +8.70 $\pm$ 1.66  \\
CIFAR-100 & LeNet     & 64.78 $\pm$ 1.08 & 69.28 $\pm$ 1.15 & +4.49 $\pm$ 1.37  \\
CIFAR-100 & ResNet-18 & 64.29 $\pm$ 0.85 & 65.84 $\pm$ 1.38 & +1.55 $\pm$ 0.67  \\
\bottomrule
\end{tabular}
\end{table}

\subsubsection{Computational Costs of FGPM}
\label{app:computational-cost}
\begin{table}[t]
    \centering
    \caption{FGM and FGPM execution times (mean $\pm$ std.) 
    across architectures.}
    \label{tab:fpgm-wall-clock}
    \small
    \begin{tabular}{llrr}
        \toprule
        Dataset & Architecture & GPM time (s) & FGPM time (s) \\
        \midrule
        CIFAR-10 & AlexNet   & $309.9 \pm 13.1$  & $292.5 \pm 13.9$ \\
        CIFAR-10 & AlexNet   & $316.1 \pm 14.9$  & $301.1 \pm 13.5$ \\
        CIFAR-10 & LeNet     & $296.8 \pm 14.8$  & $285.4 \pm 16.1$ \\
        CIFAR-10 & ResNet-18 & $1336.7 \pm 18.7$ & $677.8 \pm 17.6$ \\
        \midrule
        CIFAR-100 & AlexNet   & $420.9 \pm 23.4$  & $416.3 \pm 22.0$ \\
        CIFAR-100 & AlexNet   & $459.6 \pm 10.0$  & $462.3 \pm 6.1$ \\
        CIFAR-100 & LeNet     & $412.6 \pm 6.1$   & $429.8 \pm 19.2$ \\
        CIFAR-100 & ResNet-18 & $2193.4 \pm 32.6$ & $1815.7 \pm 26.1$ \\
        \bottomrule
    \end{tabular}
    \vspace{6pt}

    {\footnotesize Wall clock time in seconds for each continual-learning experiment (training, basis construction, projection and evaluation) at the common base threshold (e=0.97).}
\end{table}

One other practical way our method FGPM diverges from the purely geometric baseline of GPM is the requirement for backpropagation to compute the Jacobian for the task-sensitivity matrix. We report the execution times in Table \ref{tab:fpgm-wall-clock} when run on a single NVIDIA L40S. FGPM has comparable or lower total runtime in six of the eight settings. FGPM's empirical Fisher estimator uses per-example gradients of a scalar task loss. These gradients are computed directly through a vector–Jacobian product (VJP) in one backward pass, rather than calculating and storing the derivative of every output with respect to every feature. GPM avoids the additional backward pass, but for convolutional layers it expands the activations into a large matrix of overlapping image patches before identifying the directions to protect. This operation is particularly expensive for ResNet-18. FGPM can also retain fewer protected directions, making the constraints applied during subsequent training less expensive. These implementation details explain why FGPM can have a lower end-to-end runtime despite the need for backpropagation.

With respect to memory complexity, the two methods have the same asymptotic storage requirement for their final second-moment matrices. For a protected representation dimension $d$, both the GPM activation matrix and the FGPM task-gradient matrix are $d\times d$ if formed densely, requiring $O(d^2)$ memory.

\subsection{Representational comparison experiments}
\label{app:alignment}

\subsubsection{Task-sensitive vs null alignment across depth}

We compute task-conditioned CKA between all pairs of four pretrained ViT models: ViT-B/16 (\texttt{vit\_base\_patch16\_224}, torchvision), ViT-B/16 pretrained on ImageNet-21k (\texttt{vit\_base\_patch16\_224.augreg\_in21k\_ft\_in1k}, timm), DeiT3-B/16 (\texttt{deit3\_base\_patch16\_224}, timm), and ViT-L/16 (\texttt{vit\_large\_patch16\_224}, torchvision), yielding $6$ pairs.

\paragraph{Feature extraction.}
For each model, we sample 960 images from the ImageNet validation set for estimating the task-sensitivity matrix and a separate 500 images for computing alignment (held-out test split).
All images are preprocessed with standard ImageNet normalization (resize to 256, center crop to 224).
Features are extracted as the CLS token output at each transformer block via forward hooks.
Layers are matched across models by relative depth: for pairs of 12-block models, layers are matched one-to-one; for pairs involving ViT-L/16 (24 blocks), we match 8 evenly spaced layers from each model at equivalent relative positions.

\paragraph{Task-sensitivity matrix.}
At each layer $\ell$, the task-sensitivity matrix $F_\ell$ is computed using the model's own pretrained ImageNet classification head.
For each calibration image, we extract the CLS token at layer $\ell$, compute the Jacobian of the output logits (restricted to the top-1 predicted class) with respect to the CLS token through all remaining transformer blocks, the layer norm, and the classification head, and accumulate the outer product of gradients.
The matrix is symmetrized and eigendecomposed.

\paragraph{Subspace decomposition.}
We use a fixed-rank decomposition: the top-$k$ eigenvectors of $F_\ell$ define the task-sensitive subspace, and the bottom-$k$ eigenvectors define the null subspace, for $k \in \{25, 50, 100, 200, 300\}$.
Unlike an energy-threshold decomposition (which selects $k$ adaptively based on the eigenvalue spectrum), fixed-rank ensures that task and null subspaces have equal dimensionality across layers and models, making the CKA comparison fair.

\paragraph{Alignment computation.}
For each model pair at each matched layer, the held-out test representations are projected onto the task-sensitive and null subspaces (using each model's own $F_\ell$ decomposition), and linear CKA is computed separately in each subspace.
CKA is computed as $\mathrm{CKA}(X, Y) = \|X^\top Y\|_F^2 / (\|XX^\top\|_F \|YY^\top\|_F)$ on mean-centered projected representations.

\subsubsection{Task-conditioned alignment: shape vs texture}

We compute task-conditioned CKA between all pairs of four pretrained models described above. For each pair, at the penultimate layer, we train separate MLP probes (one hidden layer, 256 units, ReLU, dropout 0.1) for shape and texture classification on the Geirhos cue-conflict stimuli (960 train, 240 test; see Appendix~\ref{app:shape_texture} for stimulus details).
The task-sensitivity matrix for each probe is computed via the Jacobian through the remaining transformer blocks and the probe, using the training split.
The subspace is defined by retaining the top eigenvectors capturing 90\% of the eigenvalue mass.
Held-out test representations are projected onto each model's shape-sensitive and texture-sensitive subspaces, and CKA is computed separately.

\subsection{Task-sensitive compression changes representational geometry}
\label{app:selfalignment}

\begin{figure}[htbp]
    \centering
    \includegraphics[width=\linewidth]{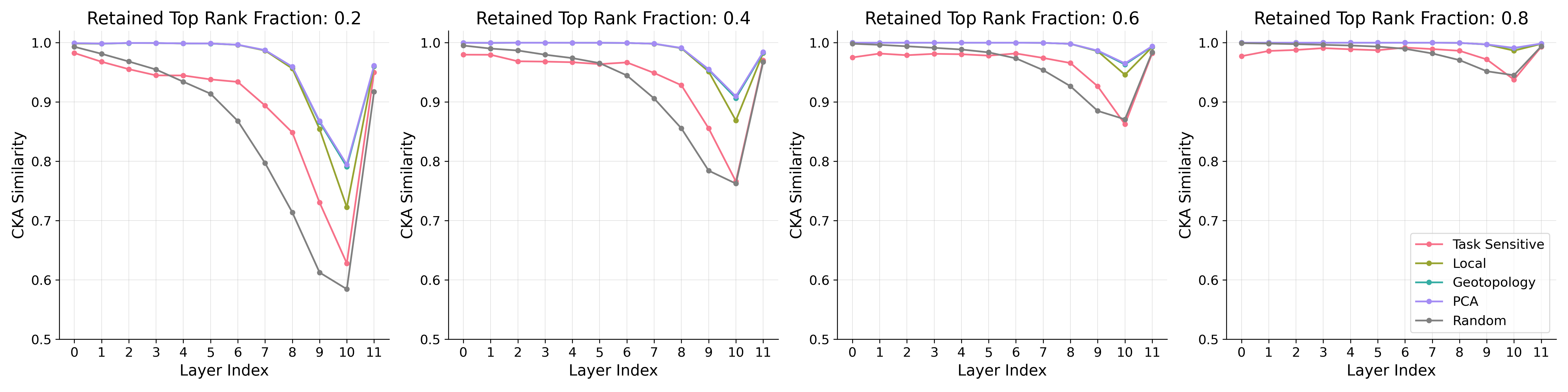}
    \caption{
        \textbf{(Layer-wise alignment Between Projected and Base Representations.}
        Centered Kernel Alignment between compressed and uncompressed representations
        at each transformer layer, shown as line plots across layers for four fixed
        retained rank fractions.
    }
    \label{fig:c1_layerwise_cka}
\end{figure}

\begin{figure}[htbp]
    \centering
    \includegraphics[width=\linewidth]{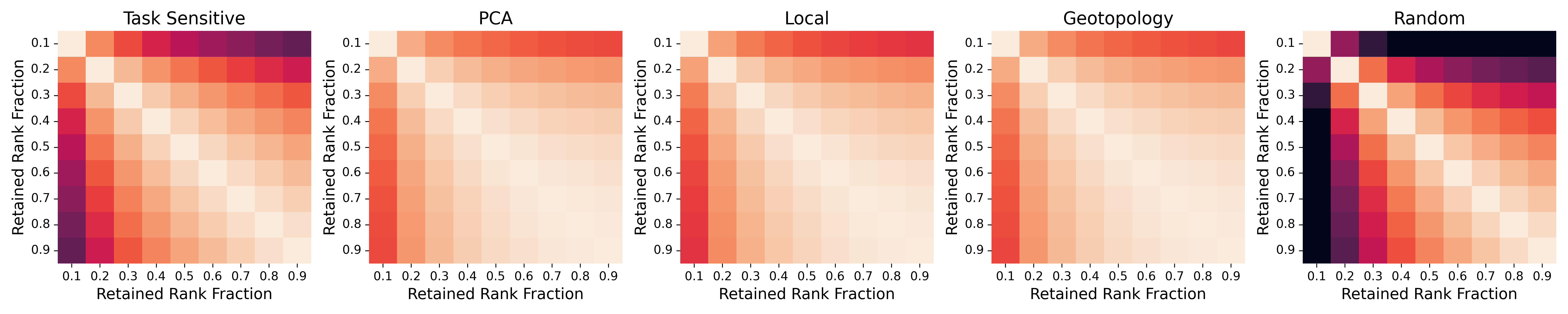}
    \caption{
        \textbf{(Projection Self-Alignment}
        CKA between representations compressed with the same method at different
        rank fractions and layers. Lighter colors are closer to CKA score of 1 and darker color is closer to 0.
    }
    \label{fig:c2_self_similarity_cka}
\end{figure}

\begin{figure}[htbp]
    \centering
    \includegraphics[width=\linewidth]{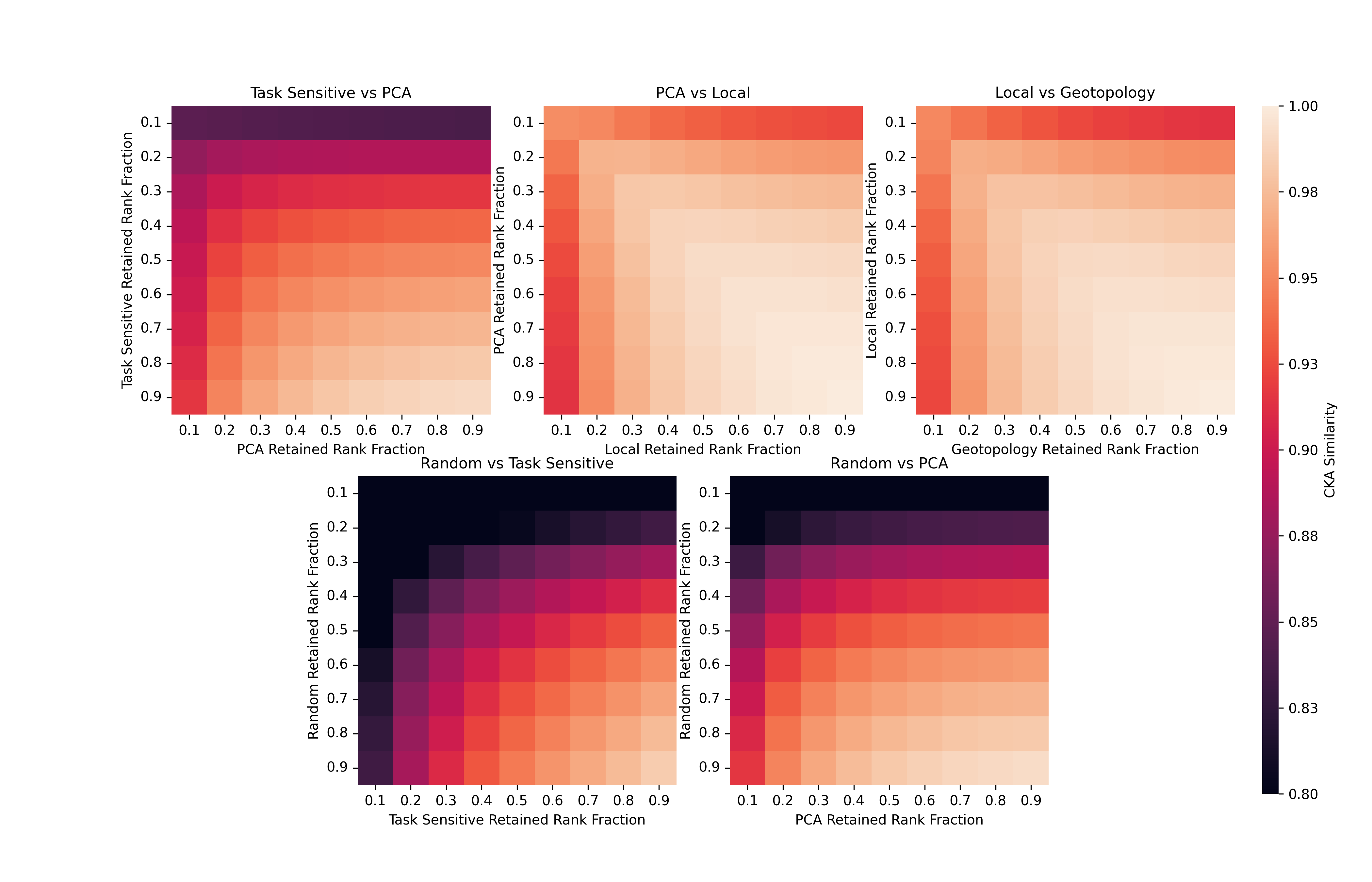}
    \caption{
        \textbf{(Cross-Projection Representational Similarity.}
        CKA between representations produced by five pairs of projection methods:
        task-sensitive vs.\ PCA, PCA vs.\ local geometry, local geometry vs.\ geotopology,
        random vs.\ task-sensitive, and random vs.\ PCA. Values are averaged across
        matched layers and shown as a function of retained rank fraction.
    }
    \label{fig:c3_cross_projection_cka}
\end{figure}

\newpage

\section*{NeurIPS Paper Checklist}

%%% END INSTRUCTIONS %%%

\begin{enumerate}

\item {\bf Claims}
    \item[] Question: Do the main claims made in the abstract and introduction accurately reflect the paper's contributions and scope?
    \item[] Answer: \answerYes{} % Replace by \answerYes{}, \answerNo{}, or \answerNA{}.
    \item[] Justification: The abstract and introduction accurately describe all four main contributions. All claims are supported by experiments in Sections 3.1--3.6. Scope is appropriately limited to pretrained vision transformers evaluated on ImageNet and CIFAR benchmarks.{}

\item {\bf Limitations}
    \item[] Question: Does the paper discuss the limitations of the work performed by the authors?
    \item[] Answer: \answerYes{} % Replace by \answerYes{}, \answerNo{}, or \answerNA{}.
    \item[] Justification: Limitations are discussed in the Discussion section.

\item {\bf Theory assumptions and proofs}
    \item[] Question: For each theoretical result, does the paper provide the full set of assumptions and a complete (and correct) proof?
    \item[] Answer:\answerNA{} % Replace by \answerYes{}, \answerNo{}, or \answerNA{}.
    \item[] Justification: The paper does not include formal theorems or proofs. The projector matrices are defined and empirically validated.

    \item {\bf Experimental result reproducibility}
    \item[] Question: Does the paper fully disclose all the information needed to reproduce the main experimental results of the paper to the extent that it affects the main claims and/or conclusions of the paper (regardless of whether the code and data are provided or not)?
    \item[] Answer: \answerYes{} % Replace by \answerYes{}, \answerNo{}, or \answerNA{}.
    \item[] Justification: All hyperparameters, model architectures, datasets, compression criteria, rank fractions, evaluation protocols, and allocation strategies are described in the main text and Appendices. The task-relevance matrix construction, eigenspace estimation procedure, and FGPM training protocol are fully specified.

\item {\bf Open access to data and code}
    \item[] Question: Does the paper provide open access to the data and code, with sufficient instructions to faithfully reproduce the main experimental results, as described in supplemental material?
    \item[] Answer: \answerYes{} % Replace by \answerYes{}, \answerNo{}, or \answerNA{}.
    \item[] Justification: Code and instructions to reproduce all experiments are available at \url{https://github.com/NeuroML-Lab/compressible-representations}. 

\item {\bf Experimental setting/details}
    \item[] Question: Does the paper specify all the training and test details (e.g., data splits, hyperparameters, how they were chosen, type of optimizer) necessary to understand the results?
    \item[] Answer:\answerYes{}
    \item[] Justification: Evaluation details are provided in Appendix A.2 (clean accuracy), Appendix A.4 (adversarial robustness), Appendix A.5 (cue-conflict subspace analysis), and Appendix A.6 (continual learning). This includes data splits, subset sizes, calibration sample counts, optimizer settings, training epochs, batch sizes, threshold selection, and rank allocation procedures.

\item {\bf Experiment statistical significance}
    \item[] Question: Does the paper report error bars suitably and correctly defined or other appropriate information about the statistical significance of the experiments?
    \item[] Answer: \answerYes{}
    \item[] Justification: Error bands (SEM across model pairs) are reported in the representational comparison figures (Figure 5). The random projection baseline includes a shaded confidence band across random seeds (Figures 2--3). The continual learning results in Table 1 are single-run evaluations following the protocol of Saha et al. (2021); variability across seeds is not reported, consistent with prior work on this benchmark.

\item {\bf Experiments compute resources}
    \item[] Question: For each experiment, does the paper provide sufficient information on the computer resources (type of compute workers, memory, time of execution) needed to reproduce the experiments?
    \item[] Answer: \answerNo{}
    \item[] Justification: Compute resources (GPU type, memory, and wall-clock time) are not reported in the current submission. Experiments were conducted on standard GPU hardware. The main computational cost is Jacobian estimation for the task-relevance matrix, which scales with the number of calibration samples and layers; all other compression criteria require only forward passes. We will include a compute table in the final version.
    
\item {\bf Code of ethics}
    \item[] Question: Does the research conducted in the paper conform, in every respect, with the NeurIPS Code of Ethics \url{https://neurips.cc/public/EthicsGuidelines}?
    \item[] Answer: \answerYes{}
    \item[] Justification: This work studies the representational structure of pretrained vision models using publicly available datasets and models. No new data was collected, no human subjects were involved, and no dual-use risks are identified. The work conforms with the NeurIPS Code of Ethics in all respects.

\item {\bf Broader impacts}
    \item[] Question: Does the paper discuss both potential positive societal impacts and negative societal impacts of the work performed?
    \item[] Answer: \answerYes{}
    \item[] Justification: This is foundational research on the structure of neural representations. Positive impacts include improved model compression, better continual learning, and more interpretable representational analysis tools. The adversarial robustness findings may contribute to understanding and mitigating adversarial vulnerability. No direct negative societal impacts are identified: the framework analyzes existing pretrained models and does not introduce new generative capabilities, surveillance tools, or decision-making systems. The work does not have an obvious path to harmful applications.
    
\item {\bf Safeguards}
    \item[] Question: Does the paper describe safeguards that have been put in place for responsible release of data or models that have a high risk for misuse (e.g., pre-trained language models, image generators, or scraped datasets)?
    \item[] Answer: \answerNA{}
    \item[] Justification: The paper does not release new models, datasets, or high-risk assets. All models and datasets used are publicly available and widely used in the research community.

\item {\bf Licenses for existing assets}
    \item[] Question: Are the creators or original owners of assets (e.g., code, data, models), used in the paper, properly credited and are the license and terms of use explicitly mentioned and properly respected?
    \item[] Answer:  \answerYes{} % Replace by \answerYes{}, \answerNo{}, or \answerNA{}.
    \item[] Justification: All models and datasets used have been attributed by citations.

\item {\bf New assets}
    \item[] Question: Are new assets introduced in the paper well documented and is the documentation provided alongside the assets?
    \item[] Answer: \answerNA{}
    \item[] Justification: The paper does not introduce new datasets or model checkpoints. 

\item {\bf Crowdsourcing and research with human subjects}
    \item[] Question: For crowdsourcing experiments and research with human subjects, does the paper include the full text of instructions given to participants and screenshots, if applicable, as well as details about compensation (if any)? 
    \item[] Answer: \answerNA{}
    \item[] Justification: The paper does not involve crowdsourcing or human subjects. All experiments use automated evaluation on standard vision benchmarks.

\item {\bf Institutional review board (IRB) approvals or equivalent for research with human subjects}
    \item[] Question: Does the paper describe potential risks incurred by study participants, whether such risks were disclosed to the subjects, and whether Institutional Review Board (IRB) approvals (or an equivalent approval/review based on the requirements of your country or institution) were obtained?
    \item[] Answer: \answerNA{}
    \item[] Justification: No human subjects research was conducted.

\item {\bf Declaration of LLM usage}
    \item[] Question: Does the paper describe the usage of LLMs if it is an important, original, or non-standard component of the core methods in this research? Note that if the LLM is used only for writing, editing, or formatting purposes and does \emph{not} impact the core methodology, scientific rigor, or originality of the research, declaration is not required.
    %this research? 
    \item[] Answer: \answerNA{} % Replace by \answerYes{}, \answerNo{}, or \answerNA{}.
    \item[] Justification: LLMs were not used as any component of the research methodology, experimental design, or analysis.

\end{enumerate}
\end{document}